\pdfoutput=1

\documentclass[11pt]{article}

\usepackage{ACL2023}

\usepackage{times}
\usepackage{latexsym}

\usepackage{float}

\usepackage{bm}
\usepackage{graphicx}
\usepackage{subfigure}
\usepackage{amssymb}
\usepackage{amsmath}
\usepackage{amsthm}
\usepackage{multirow}
\usepackage{enumitem}
\usepackage{tabularx}
\usepackage{booktabs}
\usepackage{arydshln}
\usepackage{graphicx}
\usepackage[normalem]{ulem}
\useunder{\uline}{\ul}{}

\usepackage{pifont}
\newcommand{\cmark}{\ding{51}}
\newcommand{\xmark}{\ding{55}}

\usepackage[T1]{fontenc}

\usepackage[utf8]{inputenc}

\usepackage{microtype}

\usepackage{inconsolata}

\title{Modeling User Satisfaction Dynamics in Dialogue via Hawkes Process}

\author{Fanghua Ye$^{\dagger}$ \and Zhiyuan Hu$^{\ddagger}$ \and Emine Yilmaz$^{\dagger}$ \\
        $^{\dagger}$University College London \\ $^{\ddagger}$National University of Singapore \\
        \texttt{\{fanghua.ye.19, emine.yilmaz\}@ucl.ac.uk}, \space
         \texttt{zhiyuan\_hu@u.nus.edu}}

\begin{document}
\maketitle
\begin{abstract}
  Dialogue systems have received increasing attention while automatically evaluating their performance remains challenging. User satisfaction estimation (USE) has been proposed as an alternative. It assumes that the performance of a dialogue system can be measured by user satisfaction and uses an estimator to simulate users. The effectiveness of USE depends heavily on the estimator. Existing estimators independently predict user satisfaction at each turn and ignore satisfaction dynamics across turns within a dialogue. In order to fully simulate users, it is crucial to take satisfaction dynamics into account. To fill this gap, we propose a new estimator ASAP (s\textbf{A}tisfaction e\textbf{S}timation via H\textbf{A}wkes \textbf{P}rocess) that treats user satisfaction across turns as an event sequence and employs a Hawkes process to effectively model the dynamics in this sequence. Experimental results on four benchmark dialogue datasets demonstrate that ASAP can substantially outperform state-of-the-art baseline estimators.
\end{abstract}


\section{Introduction}

Dialogue systems are playing an increasingly important role in our daily lives. They can serve as intelligent assistants to help users accomplish tasks and answer questions or as social companion bots to converse with users for entertainment \citep{ni2022recent, fu2022learning}. In recent years, the research and development of dialogue systems has made remarkable progress. However, due to the complexity of human communication, the latest dialogue systems may still fail to understand users' intents and generate inappropriate responses \citep{liang-etal-2021-turn, deng2022benefits, pan-etal-2022-user}. These deficiencies pose huge challenges to deploying dialogue systems to real-life applications, especially high-stakes ones such as finance and health. In light of this, it is crucial to evaluate the performance of dialogue systems adequately in their development phase \citep{sun2021simulating, deriu2021survey}.


Generally speaking, there are two types of evaluation methods, human evaluation and automatic evaluation \citep{deriu2021survey}. Human evaluation is fairly effective, but costly and hard to scale up. By contrast, automatic evaluation is more scalable. However, due to the ambiguity of what constitutes a high-quality dialogue, there are currently no universally accepted evaluation metrics. Existing commonly used metrics such as BLEU \citep{papineni-etal-2002-bleu} usually do not agree with human judgment. Nonetheless, user satisfaction estimation (USE) has been proposed as an alternative \citep{bodigutla2019multi, park2020large, kachuee-etal-2021-self, sun2021simulating}. USE assumes that the performance of a dialogue system can be approximated by the satisfaction of its users and simulates users' satisfaction with an estimator. In this regard, USE performs automatic evaluation and is thus scalable.

\begin{figure}[t]
  \centering
  \includegraphics[width=0.932\linewidth]{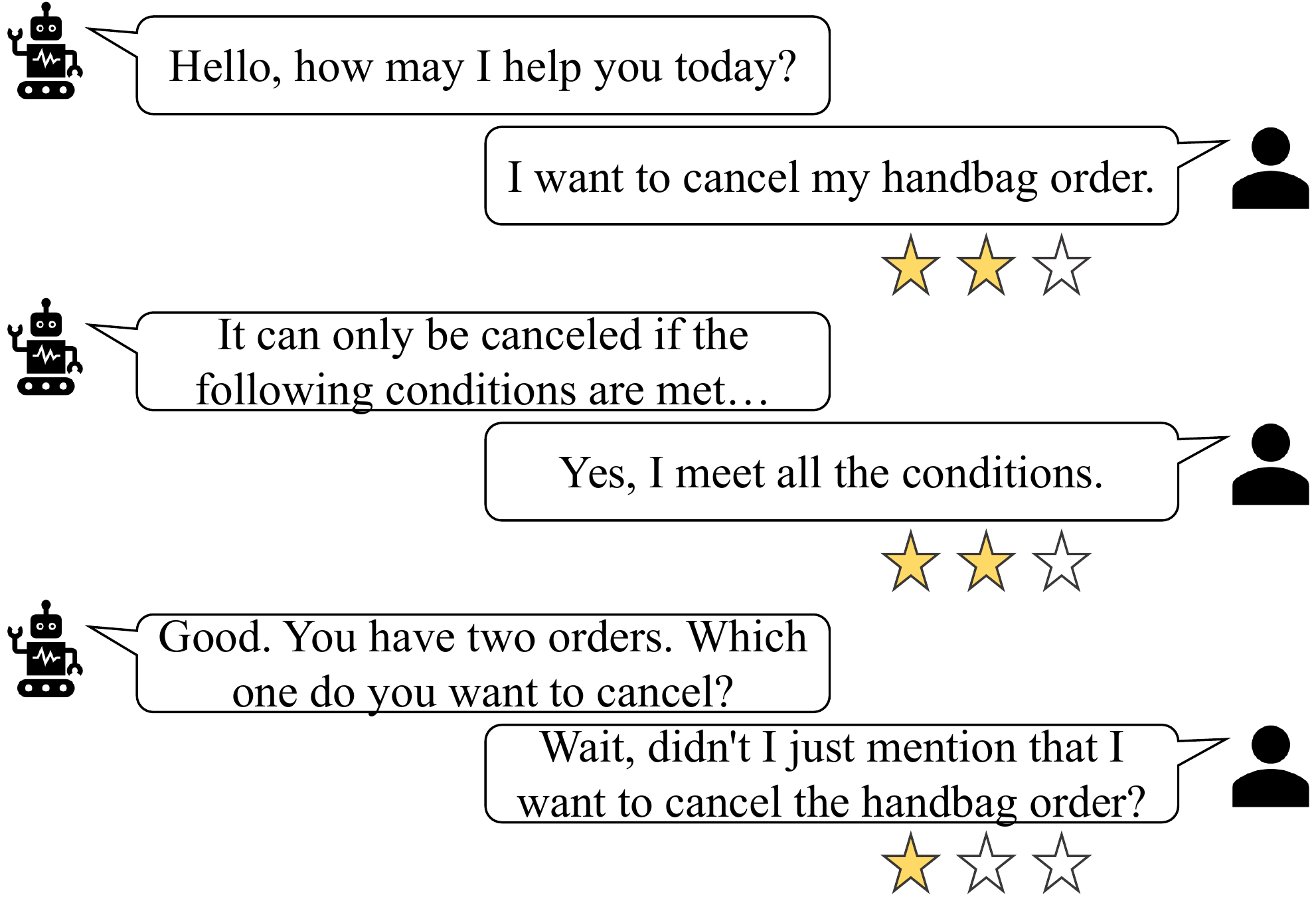}
  \caption{An example dialogue showing the dynamics of user satisfaction across different interaction turns.}
  \label{fig:example}
\end{figure}

Aside from helping developers find the defects of a dialogue system, USE also makes it possible to carry out timely human intervention for dissatisfied users and continuously optimize the system from human feedback \citep{hancock-etal-2019-learning, bodigutla-etal-2020-joint, deriu2021survey, deng2022user}. In essence, USE is a multi-class classification problem and the goal is to predict user satisfaction at each turn. Take the dialogue shown in Figure~\ref{fig:example} as an example, where user satisfaction is measured on a three-point scale. At the first two turns, the system responds appropriately. However, at the third turn, even though the response seems to be reasonable, the system asks for information that the user has already  provided at the first turn, which may lead to dissatisfaction.

As a model-based metric, the evaluation quality of USE relies heavily on the satisfaction estimator used. In order to train a robust estimator, different approaches have been proposed \citep{sun2021simulating, liang-etal-2021-turn, kachuee-etal-2021-self, pan-etal-2022-user, deng2022user}. Despite the effectiveness of these approaches, they estimate user satisfaction at each turn independently and ignore the dynamics of user satisfaction across turns within a dialogue. Given that a user's satisfaction is not only related to the current dialogue context, but may also be related to the satisfaction states at previous turns, we argue that modeling user satisfaction dynamics is valuable for training a more powerful estimator.

To achieve this, we propose ASAP (s\textbf{A}tisfaction e\textbf{S}timation via H\textbf{A}wkes \textbf{P}rocess), a novel approach that leverages Hawkes process \citep{hawkes2018hawkes} to capture the dynamics of user satisfaction. Hawkes process is  a self-exciting point process and it has been widely adopted to model sequential data such as financial transactions \citep{bacry2015hawkes} and healthcare records \citep{wang2018supervised}. In particular, we make the following contributions:
\begin{itemize}
    \item We first propose a base estimator to predict user satisfaction based solely on the dialogue context. We then incorporate a Hawkes process module to model user satisfaction dynamics by treating the satisfaction scores across turns within a dialogue as an event sequence.

    \item We propose a discrete version of the continuous Hawkes process to adapt it to the USE task and implement this module with a Transformer architecture \citep{vaswani2017attention}. 

    \item We conduct extensive experiments on four dialogue datasets. The results show that ASAP substantially outperforms baseline methods. 
\end{itemize}

\section{Problem Statement}
\label{sec:PS}

Suppose that we are provided with a dialogue session $\mathcal{X}$ containing $T$ interaction turns, denoted as $\mathcal{X} = \{(R_1, U_1), (R_2, U_2), \dots, (R_T, U_T)\}$. Each interaction turn $t$ ($1 \leq t \leq T$) consists of a response $R_t$ by the system and an utterance $U_t$ by the user. The goal of USE is to predict the user satisfaction score $s_t$ at each turn $t$ based on the dialogue context $\mathcal{X}_t = \{(R_1, U_1), (R_2, U_2),\dots,(R_t, U_t)\}$. Hence, our task is to learn an estimator $\mathcal{E}: \mathcal{X}_t \rightarrow s_t$ that can accurately estimate the user's satisfaction throughout the entire dialogue session.

Previous studies have shown that adding user action recognition (UAR) as an auxiliary task can facilitate the training of a stronger satisfaction estimator \citep{sun2021simulating, deng2022user}. When user action labels are available, our task shifts to learning an estimator $\mathcal{E}': \mathcal{X}_t \rightarrow (s_t, a_t)$ that predicts user satisfaction and user action simultaneously. Here, $a_t$ denotes the user action at turn $t$.

\section{Method}

\begin{figure}[t]
  \centering
  \includegraphics[width=1.0\linewidth]{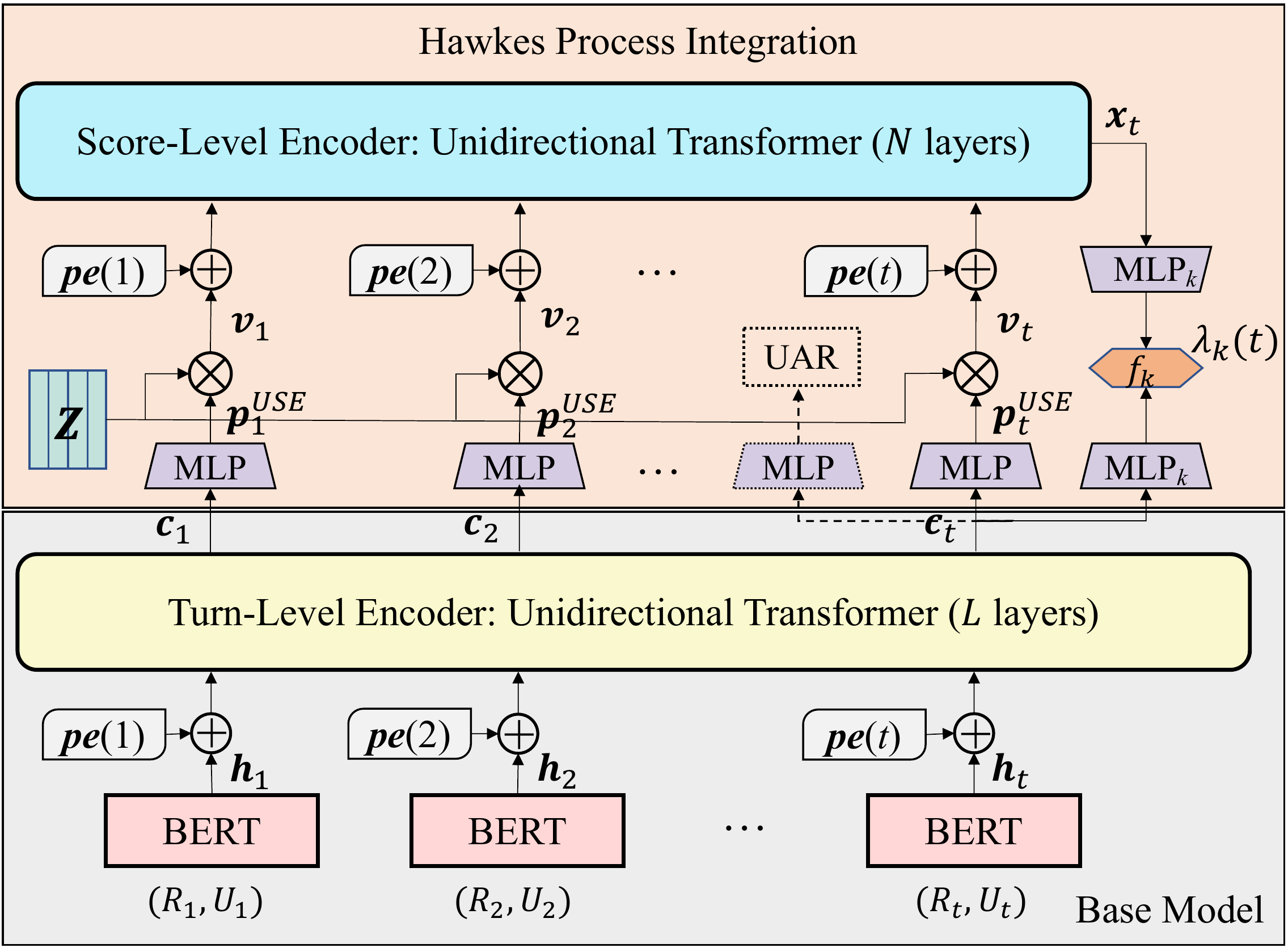}
  \caption{The architecture of our model ASAP. It consists of a base estimator module and a Hawkes process integration module. Both modules leverage positional encodings to retain temporal information. Note that a single BERT model is shared by all turns and the (optional) UAR component is depicted in dashed lines.}
  \label{fig:framework}
\end{figure}

In this section, we first describe how to build a base USE model leveraging only the dialogue context and without modeling the dynamics of user satisfaction. Then, we extend this model by integrating the Hawkes process to capture the dynamic changes of user satisfaction across dialogue turns. The overall model architecture is illustrated in Figure~\ref{fig:framework}.

\subsection{Base Satisfaction Estimator}
\label{sec:basees}

Similar to \citet{deng2022user}, we utilize a hierarchical transformer architecture to encode the dialogue context $\mathcal{X}_t$ into contextual semantic representations. A hierarchical architecture enables us to handle long dialogues.  This architecture consists of a token-level encoder and a turn-level encoder.


\subsubsection{Token-Level Encoder}

The token-level encoder takes as input the concatenation of the system response $R_t$ and user utterance $U_t$ at each turn $t$ and yields a single vector $\bm{h}_t$ as their semantic vector representation. To be specific, we adopt the pre-trained language model BERT \citep{devlin-etal-2019-bert} to encode each $(R_t, U_t)$ pair:
\begin{equation}
        \bm{h}_t = \mathtt{BERT}([CLS]  R_t  [SEP]  U_t  [SEP]).
\end{equation}

\subsubsection{Turn-Level Encoder}
\label{sec:tlencoder}

The token-level encoder can only capture the contextual information within each turn. In order to capture the global contextual information across turns, we develop a turn-level encoder that takes the semantic representations $\{\bm{h}_1, \bm{h}_2, \dots, \bm{h}_t\}$ of all turns in the dialogue context $\mathcal{X}_t$ as input. We implement this encoder as a unidirectional Transformer encoder with $L$ layers. Similar to the standard Transformer encoder layer \citep{vaswani2017attention}, each layer includes two sub-layers. The first sub-layer is a masked multi-head attention module ($\mathtt{MultiHead}$). The second sub-layer is a position-wise feed-forward network which is composed of two linear transformations with a ReLU activation in between ($\mathtt{FFN}$). 

Formally, each layer of the turn-level encoder operates as follows:
\begin{align}
    &\bm{H}^{(0)} = [\bm{h}_1 + \bm{pe}(1), \dots, \bm{h}_t+\bm{pe}(t)], \\
    &\bm{H}^* = \mathtt{MultiHead}(\bm{H}^{(l)}, \bm{H}^{(l)}, \bm{H}^{(l)}) , \\
    &\bm{H}^{(l+1)} = \mathtt{FFN}(\bm{H}^* + \bm{H}^{(l)}) + \bm{H}^* + \bm{H}^{(l)}, 
\end{align}
where $\bm{H}^{(0)}$ is the input of the first layer, in which we add positional encodings $\bm{pe}(\cdot)$ to retain the turn order information. We calculate $\bm{pe}(\cdot)$ in the same way as \citet{vaswani2017attention}. $\bm{H}^{(L)} = [\bm{c}_1, \dots, \bm{c}_t]$ is the output of the last layer with $\bm{c}_t$ denoting the final contextualized representation of the $t$-th turn. Notice that layer normalization \citep{ba2016layer} 
is omitted in the formulae above for simplicity.

\subsubsection{Satisfaction Estimation}

After acquiring the contextual representation $\bm{c}_t$, we can readily compute the probability distribution of user satisfaction at turn $t$ by applying an MLP network \citep{rumelhart1986learning} with softmax normalization to $\bm{c}_t$, as shown below:
\begin{equation}
\label{eq:prob}
    \bm{p}^{USE}_t = \mathtt{softmax}(\mathtt{MLP}(\bm{c}_t)),
\end{equation}
where $\bm{p}^{USE}_t \in \mathbb{R}^K$, and $K$ is the number of satisfaction classes. The class with the highest probability is selected as the prediction.

\subsection{Hawkes Process Integration}

\subsubsection{Preliminaries on Hawkes Process}

The Hawkes process is a self-exciting point process. It models the self-excitation of events having the same type and the mutual excitation of events with different types in an additive way. A Hawkes process is characterized by its conditional intensity function, which is defined as:
\begin{equation}
    \label{eq:baseintensity}
    \lambda(t) = \mu(t) + \sum_{t_i: t_i < t} \psi (t - t_i).
\end{equation}
Here, $t_i$ denotes the occurrence time of a past event, $\mu(t) > 0$ is the background intensity or base intensity, and $\psi(\cdot) \geq 0$ is a pre-specified triggering kernel function. Typically, $\psi(\cdot)$ is chosen to be a time-decaying function (e.g., the exponential function $\exp(-t)$), indicating that the impacts of past events on the current event decrease through time. 

While being able to model the influence of past events, the formulation in Eq.~\eqref{eq:baseintensity} is too simple to capture the complicated dynamics of many real-life event sequences. For example, it assumes that each of the past events has a positive effect on the occurrence of the current event, which can be unrealistic in numerous complex scenarios. To improve its capability, neural Hawkes process models have been devised \citep{mei2017neural, xiao2017modeling}. These models generalize the standard Hawkes process by parameterizing its intensity function with recurrent neural networks (RNNs) such as LSTM \citep{hochreiter1996lstm}. More concretely, the new intensity function is calculated in the following way:
\begin{equation}
    \label{eq:dec}
    \lambda(t) = \sum^M_{m=1} \lambda_m(t) = \sum^M_{m=1}f_m(\bm{w}^T_m\bm{x}_t),
\end{equation}
where $M$ is the total number of event types, $\bm{x}_t$ is the hidden state of the event sequence, and $\bm{w}_m$ is a parameter vector that converts $\bm{x}_t$ to a scalar. $f_m(\cdot)$ is the softplus function with a ``softness'' parameter $\beta_m$, i.e., $f_m(y) = \beta_m \log(1+\exp(y/\beta_m))$. It guarantees that the intensity $\lambda(t)$ is always positive. In addition to the stronger expressiveness, this formulation of the intensity function has another advantage in that the probability of each event type $m$ can be simply calculated as $\lambda_m(t)/\lambda(t)$.

The RNNs-based Hawkes process models inherit the intrinsic weaknesses of RNNs. Inspired by the superiority of Transformers over RNNs in dealing with sequential data, several Transformer Hawkes process models have been proposed recently \citep{zuo2020transformer, zhang2020self, zhou2022neural}. For these models, one representative definition of the type-specific intensity function $\lambda_m(t)$ takes the form \citep{zuo2020transformer}:
\begin{equation}
    \label{eq:hawkesTF}
    \lambda_m(t) = f_m\left(\alpha_m \frac{t-t_i}{t_i} + \bm{w}^T_m\bm{x}_{t_i} + b_m\right).
\end{equation}
In Eq.~\eqref{eq:hawkesTF}, $b_m$ represents the base intensity and $\alpha_m$ is introduced to modulate the importance of time interpolation. This interpolation enables $\lambda_m(t)$ to be continuous over time. The overall intensity function $\lambda(t)$ is still defined as $\lambda(t)=\sum^M_{m=1}\lambda_m(t)$.

\subsubsection{Adapting Hawkes Process for Satisfaction Estimation}

Intuitively, the user satisfaction scores across turns within a dialogue can be regarded as an event sequence and each score corresponds to one type of event. Therefore, it is a natural fit to adopt Hawkes process to model the dynamics of user satisfaction. However, it is infeasible to apply the standard Hawkes process or its neural variants mentioned above directly. This is because these Hawkes processes are continuous in time, i.e., the domain of their intensity function $\lambda(t)$ is the interval $(0, T]$. A continuous Hawkes process models both \textit{what} the next event type will be and \textit{when} the next event will happen. By comparison, the satisfaction score sequence in our case is \textit{discrete} in time. We only need to predict the next event type (i.e., the satisfaction score) and there is no need to predict when it will happen as we estimate user satisfaction at every turn. This difference inspires us to design a discrete version of the Hawkes process.

It is worth emphasizing that one satisfaction prediction is supposed to be made at every dialogue turn, meaning that one event regardless of its type will certainly happen at each turn. To achieve this, we constrain the intensity function $\lambda(t)$ to always take the value 1. Furthermore, following Eq.~\eqref{eq:dec}, $\lambda(t)$ is decomposed into:
\begin{equation}
\label{eq:ourintensity}
\begin{aligned}
    &\lambda(t) = \sum^K_{k=1} \lambda_k(t) = 1, ~ t \in \{1, 2, \dots, T\}, \\
    &\mbox{s.t.} ~~ \lambda_k(t) > 0, ~~  \forall k =1, 2, \dots, K. 
\end{aligned}
\end{equation}
Recall that $K$ represents the number of satisfaction classes. Due to $\lambda(t)=1$, $\lambda_k(t)$ can be regarded as the probability that event type $k$ happens (i.e., the satisfaction score is $k$). In Eq.~\eqref{eq:ourintensity}, $\lambda(t)$ is defined on the discrete domain $\{1, 2, \dots, T\}$ rather than the continuous interval $(0, T]$.

We propose to calculate each $\lambda_k(t)$ by the following formula:
\begin{equation}
    \label{eq:kkkkkkkk}
    \lambda_k(t) = \frac{\mathtt{exp}\big(f_k(\mathtt{MLP}_k(\bm{c}_t) + \mathtt{MLP}_k(\bm{x}_t))\big)}{\sum^K_{j=1} \mathtt{exp}\big(f_j(\mathtt{MLP}_j(\bm{c}_t) + \mathtt{MLP}_j(\bm{x}_t))\big)},
\end{equation}
where the term associated with $\bm{c}_t$ characterizes the contribution of the dialogue context $\mathcal{X}_t$ to the intensity (i.e., base intensity) and the term corresponding to $\bm{x}_t$ reveals the contribution of the satisfaction sequence. Different from Eqs.~\eqref{eq:dec} and \eqref{eq:hawkesTF}, we perform non-linear rather than linear transformations to convert both $\bm{c}_t$ and $\bm{x}_t$ into scalars using MLP networks. Note that $f_k(\cdot)$ is the softplus function.

Next, we describe how to compute $\bm{x}_t$, the hidden state of the satisfaction score sequence. Given the strong capability of Transformer Hawkes process models, we choose to employ a Transformer architecture (named \textbf{score-level encoder}) to compute $\bm{x}_t$. In particular, we adopt a unidirectional Transformer with $N$ layers. Same as the turn-level encoder (refer to \S\ref{sec:tlencoder}), each layer contains two sub-layers, the multi-head attention sub-layer and the position-wise feed-forward sub-layer. 

The input to its first layer is the satisfaction score sequence. To convert this sequence into vector representations, we introduce an embedding matrix $\bm{Z} \in \mathbb{R}^{d \times K}$ whose $k$-th column is a $d$-dimensional embedding for satisfaction class $k$. In principle, if we have the ground-truth score $s_t$ for turn $t$, we can calculate the embedding vector of this turn as $\bm{Z}\bm{e}_{s_t}$, where $\bm{e}_{s_t}$ is the one-hot encoding of score $s_t$. In practice, however, we need to predict the satisfaction scores for all turns. Let $\hat{s}_t$ be the predicted score of turn $t$ and $\bm{Z}\bm{e}_{\hat{s}_t}$ the corresponding embedding vector. Then, we can feed $[\bm{Z}\bm{e}_{\hat{s}_1}, \dots, \bm{Z}\bm{e}_{\hat{s}_t}]$ to the score-level encoder to learn the dynamics of user satisfaction up to turn $t$ and to obtain $\bm{x}_t$. This approach, albeit straightforward, has a severe limitation that there is no feedback from the score-level encoder to help train the base model because the gradients from the score-level encoder cannot be back-propagated to the base model. To overcome this limitation, we take the probability distribution of satisfaction classes $\bm{p}^{USE}_t$, as shown in Eq.~\eqref{eq:prob}, as the predicted ``soft'' score. Then, the embedding vector of turn $t$ is computed by:
\begin{equation}
    \bm{v}_t = \bm{Z} \bm{p}^{USE}_t.
\end{equation}
It can be seen that $\bm{v}_t$ is a weighted sum of the embeddings of all satisfaction scores and the weights are the predicted probability by the base model.

Based on $\bm{v}_t$, the score-level encoder functions as follows to yield $\bm{x}_t$:
\begin{align}
    &\bm{V}^{(0)} = [\bm{v}_1 + \bm{pe}(1), \dots, \bm{v}_t+\bm{pe}(t)], \\
    &\bm{V}^* = \mathtt{MultiHead}(\bm{V}^{(n)}, \bm{V}^{(n)}, \bm{V}^{(n)}) , \\
    &\bm{V}^{(n+1)} = \mathtt{FFN}(\bm{V}^* + \bm{V}^{(n)}) + \bm{V}^* + \bm{V}^{(n)}. 
\end{align}
Similar to the turn-level encoder, we add positional encodings into the input of the first layer $\bm{V}^{(0)}$ to retain the temporal information. The output of the last layer is symbolized as  $\bm{V}^{(N)} = [\bm{x}_1, \dots, \bm{x}_t]$.

\subsection{Training Objective}

We employ the cross-entropy loss as our training objective. 
Recall that $\lambda_k(t)$ represents the probability of the satisfaction score being $k$ at turn $t$. Thus, the training objective of USE is defined as:
\begin{equation}
    \mathcal{L}_{USE} = - \mathtt{log} ~p(s_t|\mathcal{X}_t) = - \mathtt{log} ~\lambda_{s_t} (t),
\end{equation}
where $s_t$ is the ground-truth satisfaction label.

As stated in \S\ref{sec:PS}, adding UAR as an auxiliary task has the potential to help us train a more powerful satisfaction estimator. Even though the proposed Transformer Hawkes process model is expected to improve the performance of USE significantly, it is still meaningful to study if adding this auxiliary task can further improve the performance. To this end, we leverage an MLP network with softmax normalization on top of the turn-level encoder to calculate the probability distribution of user action when the ground-truth labels are provided:
\begin{equation}
    \bm{p}^{UAR}_t = \mathtt{softmax}(\mathtt{MLP}(\bm{c}_t)).
\end{equation}
Let $\bm{p}^{UAR}_{t, a_t}$ be the probability corresponding to the ground-truth action label $a_t$ at turn $t$. The training objective of UAR is then defined as:
\begin{equation}
    \mathcal{L}_{UAR} = - \mathtt{log} ~ p(a_t|\mathcal{X}_t) = - \mathtt{log} ~ \bm{p}^{UAR}_{t, a_t}.
\end{equation}

We jointly optimize USE and UAR by minimizing the following loss:
\begin{equation}
    \mathcal{L}_{joint} = \mathcal{L}_{USE} + \gamma \mathcal{L}_{UAR}.
\end{equation}
Here, $\gamma$ is a hyper-parameter that controls the contribution of the UAR task.

\section{Experimental Setup}

In what follows, we detail the experimental setup. 

\subsection{Datasets \& Evaluation Metrics}

We conduct our experiments on four publicly available dialogue datasets, including MultiWOZ 2.1 (MWOZ) \citep{eric-etal-2020-multiwoz}, Schema Guided Dialogue (SGD) \citep{rastogi2020towards}, JDDC \citep{chen-etal-2020-jddc}, and Recommendation Dialogues (ReDial) \citep{li2018towards}. In particular, we perform evaluations on the subsets of these datasets with user satisfaction annotations, which are provided on a five-point scale by \citet{sun2021simulating}. Following existing works \citep{deng2022user, pan-etal-2022-user}, the satisfaction annotations are mapped into three-class labels \{\textit{dissatisfied, neutral, satisfied}\}. MWOZ, SGD, and ReDial are in English and all contain 1000 dialogues. While JDDC is a Chinese dataset and has 3300 dialogues. Except for ReDial, all the other three datasets have user action labels. The number of action types in MWOZ, SGD, and JDDC is 21, 12, and 236, respectively. For more details about these datasets, refer to \citet{sun2021simulating}.

Following previous studies \citep{cai2020predicting, song-etal-2019-using, choi2019offline, deng2022user}, we use Accuracy (\textbf{Acc}) and Macro-averaged Precision (\textbf{P}), Recall (\textbf{R}), and F1 score (\textbf{F1}) as the evaluation metrics in our experiments.


\subsection{Baseline Methods}

We compare our proposed method ASAP with several state-of-the-art baseline methods in both single-task learning and multi-task learning settings.\footnote{The implementation of ASAP is available at \url{https://github.com/smartyfh/ASAP}.} 

In the single-task learning setting, we only consider the USE task and the selected baselines are:
\begin{itemize}[label={}, leftmargin=*]
  \item \textbf{HiGRU} \citep{jiao-etal-2019-higru}, which utilizes a hierarchical GRU structure \citep{cho-etal-2014-properties} to encode the dialogue context. 
  
  \item \textbf{HAN} \citep{yang-etal-2016-hierarchical}, which adds a two-level attention mechanism to HiGRU. 
  
  \item \textbf{BERT} \citep{devlin-etal-2019-bert}, which concatenates all the utterances in the dialogue context as a flat sequence. In addition, long sequences with more than 512 tokens are truncated automatically.
  
  \item \textbf{USDA} \citep{deng2022user}, which leverages a hierarchical Transformer architecture to encode the dialogue context.
\end{itemize}

\begin{table*}[t]
\centering
\resizebox{\textwidth}{!}{%
\begin{tabular}{lccccccccccccc}
\hline
\multirow{2}{*}{\textbf{Models}} & \multirow{2}{*}{\textbf{IDPT}} & \multicolumn{4}{c}{\textbf{MWOZ}} & \multicolumn{4}{c}{\textbf{SGD}} & \multicolumn{4}{c}{\textbf{JDDC}} \\ 
\cmidrule(lr){3-6} \cmidrule(lr){7-10} \cmidrule(lr){11-14}
 &  & \textbf{Acc} & \textbf{P} & \textbf{R} & \textbf{F1} & \textbf{Acc} & \textbf{P} & \textbf{R} & \textbf{F1} & \textbf{Acc} & \textbf{P} & \textbf{R} & \textbf{F1} \\ \hline
HiGRU & \xmark & 44.6 & 43.7 & 44.3 & 43.7 & 50.0 & 47.3 & 48.4 & 47.5 & 59.7 & 57.3 & 50.4 & 52.0 \\
HAN & \xmark & 39.0 & 37.1 & 37.1 & 36.8 & 47.7 & 47.1 & 44.8 & 44.9 & 58.4 & 54.2 & 50.1 & 51.2 \\
BERT & \xmark & 46.1 & 45.5 & 47.4 & 45.9 & 56.2 & 55.0 & 53.7 & 53.7 & 60.4 & 59.8 & 58.8 & 59.5 \\ \hdashline
USDA & \cmark & 49.9 & 49.2 & 49.0 & 48.9 & 61.4 & 60.1 & 55.7 & 57.0 & 61.8 & 62.8 & 63.7 & 61.7 \\
USDA$^{\dagger}$ & \cmark & 47.0 & 45.4 & 45.6 & 45.4 & 60.2 & 60.1 & 57.6 & 58.2 & 60.2 & 60.9 & 66.0 & 61.0 \\
\textbf{ASAP} & \xmark & \textbf{56.3}$^{\ddagger}$ & \textbf{55.1}$^{\ddagger}$ & \textbf{55.4}$^{\ddagger}$ & \textbf{55.0}$^{\ddagger}$ & \textbf{64.5}$^{\ddagger}$ & \textbf{62.4}$^{\ddagger}$ & \textbf{61.9}$^{\ddagger}$ & \textbf{62.1}$^{\ddagger}$ & \textbf{65.4}$^{\ddagger}$ & \textbf{64.2}$^{\ddagger}$ & \textbf{68.5}$^{\ddagger}$ & \textbf{65.3}$^{\ddagger}$ \\ \hline
\end{tabular}
}
\caption{Single-task performance comparison. $\dagger$ indicates our reproduced results. $\ddagger$ means significant performance improvements over USDA (measured by a paired $t$-test at $p < 0.05$). IDPT is short for in-domain pre-training.}
\label{tab:main-single}
\end{table*}

\begin{table*}[t]
\centering
\resizebox{\textwidth}{!}{%
\begin{tabular}{lccccccccccccc}
\hline
\multirow{2}{*}{\textbf{Models}} & \multirow{2}{*}{\textbf{IDPT}} & \multicolumn{4}{c}{\textbf{MWOZ}} & \multicolumn{4}{c}{\textbf{SGD}} & \multicolumn{4}{c}{\textbf{JDDC}} \\ 
\cmidrule(lr){3-6} \cmidrule(lr){7-10} \cmidrule(lr){11-14}
 &  & \textbf{Acc} & \textbf{P} & \textbf{R} & \textbf{F1} & \textbf{Acc} & \textbf{P} & \textbf{R} & \textbf{F1} & \textbf{Acc} & \textbf{P} & \textbf{R} & \textbf{F1} \\ \hline
JointDAS & \xmark & 44.8 & 42.7 & 43.0 & 42.8 & 55.7 & 52.2 & 52.4 & 52.3 & 58.5 & 55.8 & 55.1 & 55.4 \\
Co-GAT & \xmark & 46.8 & 44.8 & 44.0 & 44.2 & 56.8 & 55.9 & 55.9 & 55.6 & 60.2 & 59.3 & 62.9 & 60.1 \\
~~~+BERT & \xmark & 47.0 & 46.4 & 47.2 & 46.3 & 58.6 & 55.2 & 55.7 & 55.5 & 60.6 & 60.6 & 63.7 & 61.0 \\
JointUSE & \xmark & 47.6 & 44.6 & 44.9 & 44.7 & 57.4 & 55.0 & 54.8 & 54.7 & 58.3 & 56.6 & 58.7 & 57.2 \\
~~~+BERT & \xmark & 48.9 & 47.2 & 48.0 & 47.3 & 59.0 & 57.4 & 57.1 & 57.3 & 63.8 & 60.8 & 58.6 & 59.2 \\ \hdashline
USDA & \cmark & 52.9 & 51.8 & 50.2 & 50.6 & 62.5 & 60.3 & 59.9 & 60.1 & 63.0 & 61.4 & 65.7 & 62.6 \\
USDA$^{\dagger}$ & \cmark & 49.2 & 47.7 & 48.3 & 47.9 & 61.3 & 58.4 & 59.5 & 58.8 & 61.6 & 60.0 & 62.3 & 60.7 \\
\textbf{ASAP} & \xmark & \textbf{58.1}$^{\ddagger}$ & \textbf{58.1}$^{\ddagger}$ & \textbf{54.7}$^{\ddagger}$ & \textbf{55.6}$^{\ddagger}$ & \textbf{64.8}$^{\ddagger}$ & \textbf{63.0}$^{\ddagger}$ & \textbf{62.3}$^{\ddagger}$ & \textbf{62.6}$^{\ddagger}$ & \textbf{64.1}$^{\ddagger}$ & \textbf{62.6}$^{\ddagger}$ & \textbf{67.3}$^{\ddagger}$ & \textbf{63.9}$^{\ddagger}$ \\ \hline
\end{tabular}
}
\caption{Multi-task performance comparison. $\dagger$ indicates our reproduced results. $\ddagger$ means significant performance improvements over USDA (measured by a paired $t$-test at $p < 0.05$). IDPT is short for in-domain pre-training.}
\label{tab:main-multi}
\end{table*}

In the multi-task learning setting, we consider both the USE task and UAR task. And we compare ASAP to the following baseline methods:
\begin{itemize}[label={}, leftmargin=*]
  \item \textbf{JointDAS} \citep{cerisara-etal-2018-multi}, which jointly performs UAR and sentiment classification. We replace sentiment classification with the USE task. 
  
  \item \textbf{Co-GAT} \citep{qin2021co}, which leverages graph attention networks \citep{velivckovic2017graph} to perform UAR and sentiment classification. We also replace sentiment classification with the USE task. 
  
  \item \textbf{JointUSE} \citep{bodigutla-etal-2020-joint}, which adopts LSTM \citep{hochreiter1996lstm} for learning temporal dependencies across turns. 
  
  \item \textbf{USDA} \citep{deng2022user}, which uses CRF \citep{lafferty2001conditional} to model the sequential dynamics of user actions to facilitate USE.
\end{itemize}

Our method ASAP is closely related to USDA. The main difference is that USDA focuses on modeling user action dynamics while ASAP focuses on modeling user satisfaction dynamics. Given that user action labels may not be available in practice, our method is more applicable.

\section{Experimental Results}

\subsection{Baseline Comparison}


\textbf{Single-Task Learning.} The results of single-task learning are summarized in Table~\ref{tab:main-single} and Table \ref{tab:main-red}. It can be observed that our proposed method ASAP consistently outperforms all baseline methods on all datasets. Notably, ASAP shows substantially higher performance than USDA over all four metrics even though USDA conducts in-domain pre-training to strengthen its capability of representation learning. For example, ASAP achieves $9.6\%$, $3.9\%$, $4.3\%$, and $6.9\%$ F1 score improvements on MWOZ, SGD, JDDC, and ReDial, respectively.

\begin{table}[t]
\centering
\setlength{\tabcolsep}{2.3mm}
\begin{tabular}{lcccc}
\hline
\textbf{Models} & \textbf{Acc} & \textbf{P} & \textbf{R} & \textbf{F1} \\ \hline
HiGRU & 46.1 & 44.4 & 44.0 & 43.5 \\
HAN & 46.3 & 40.0 & 40.3 & 40.0 \\
BERT & 53.6 & 50.5 & 51.3 & 50.0 \\
\hdashline
USDA & 57.3 & 54.3 & 52.9 & 53.4 \\
USDA$^{\dagger}$ & 58.1 & 55.7 & 54.5 & 54.7 \\
\textbf{ASAP} & \textbf{66.0}$^{\ddagger}$ & \textbf{62.0}$^{\ddagger}$ & \textbf{61.3}$^{\ddagger}$ & \textbf{61.6}$^{\ddagger}$ \\ \hline
\end{tabular}

\caption{Performance comparison on ReDial.}
\label{tab:main-red}
\end{table}

\begin{figure*}[t!]
	\centering
	\subfigure[{SGD (single-task)}]{
		\includegraphics[width=0.65\columnwidth]{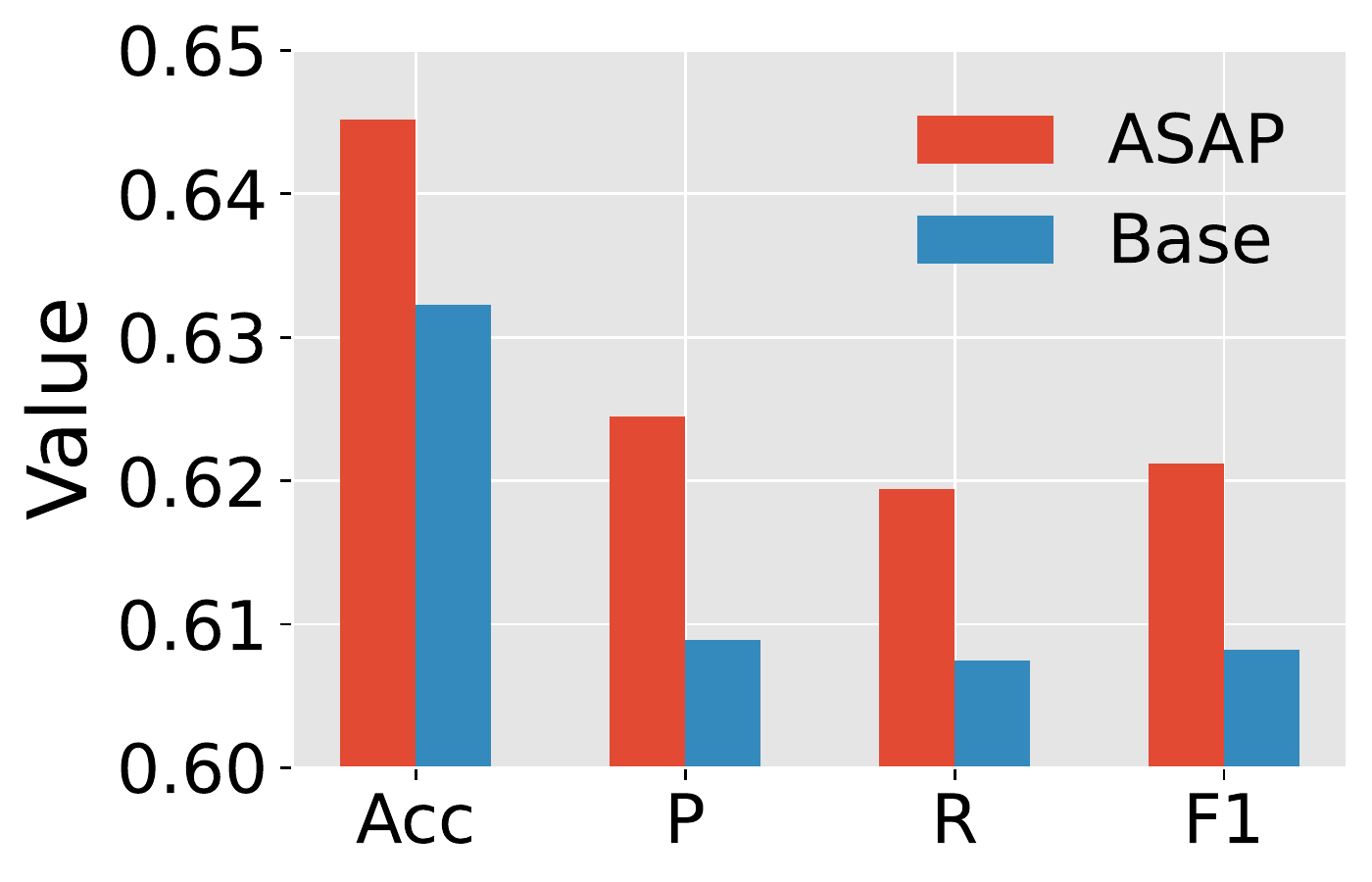}
	} \hfill
	\subfigure[{SGD (multi-task)}]{
		\includegraphics[width=0.65\columnwidth]{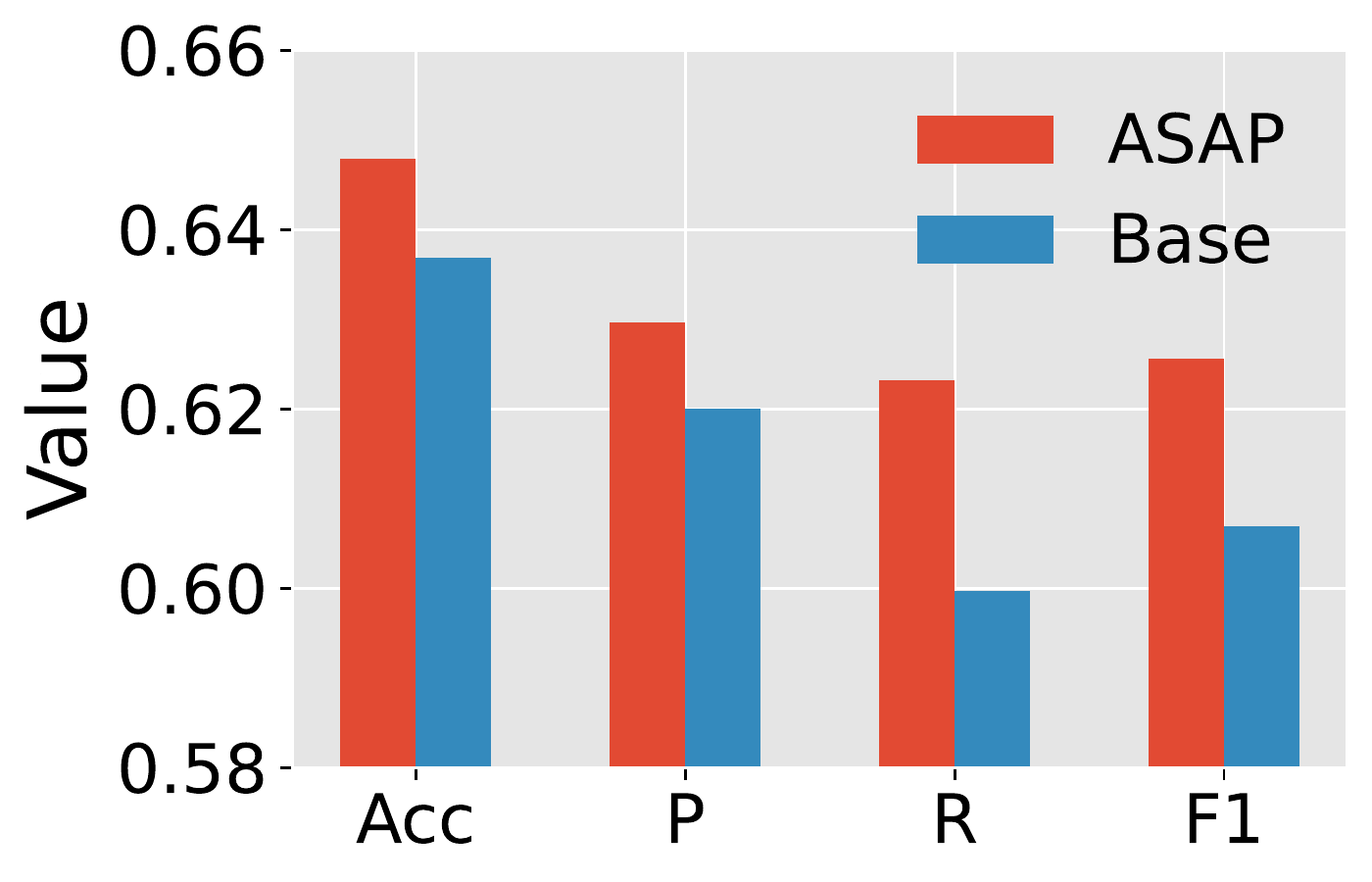}
	} \hfill
        \subfigure[{ReDial (single-task)}]{
		\includegraphics[width=0.65\columnwidth]{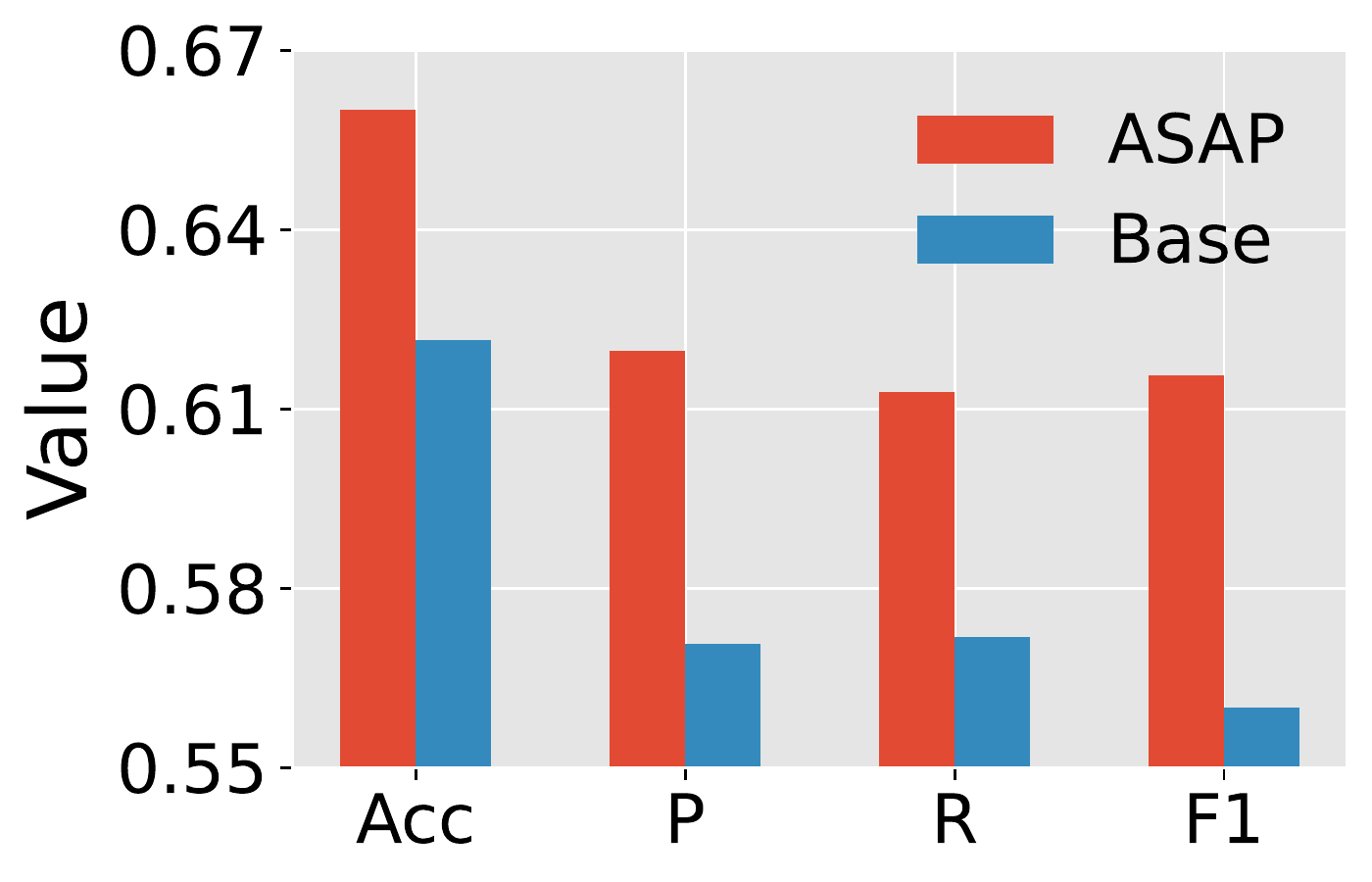}
	}
	\vspace*{-0.2cm}
	\caption{Performance comparison between ASAP and the proposed base satisfaction estimator on SGD and ReDial.}
	\label{fig:base_asap}
	\vspace*{-0.15cm}
\end{figure*}

\noindent \textbf{Multi-Task Learning.} The results of USE in the multi-task learning setting are reported in Table \ref{tab:main-multi}. For Co-GAT and JointUSE, we include results when the BERT model is leveraged. It can also be observed that the performance of ASAP is consistently higher than all baseline methods over all four metrics. For example, when compared to USDA, we observe that ASAP achieves $7.7\%$, $3.8\%$, and $3.2\%$ absolute point improvements in terms of F1 score on MWOZ, SGD, and JDDC, respectively.


\noindent \textbf{Single-Task Learning vs. Multi-Task Learning.} From Tables \ref{tab:main-single} and \ref{tab:main-multi}, we can find that ASAP tends to perform better in the multi-task learning setting on MWOZ and SGD. This indicates that adding UAR as an auxiliary task is beneficial for improving performance. However, it is worth noting that the performance gain is relatively low. To be specific, the improvements of F1 score on MWOZ and SGD are merely $0.6\%$ and $0.5\%$, respectively. Besides, on the JDDC dataset, ASAP even performs worse in the multi-task learning setting due to the large number (i.e., 236) of action types. The strong performance of ASAP in the single-task learning setting verifies the significance of modeling user satisfaction dynamics, especially considering that it is costly to collect user action labels.

In summary, our proposed method ASAP is able to outperform baseline methods in both the single-task learning setting and multi-task learning setting. Most importantly, it can achieve highly competitive performance in the single-task learning setting.

\subsection{Effectiveness of Hawkes Process Integration}

The above results have demonstrated the effectiveness of our method ASAP as a whole. However, it is unclear how much the Hawkes process integration module (i.e., the satisfaction dynamics modeling module) contributes to the overall performance. To better understand the effectiveness of this module, we conduct an ablation study where we compare the performance of ASAP with that of the base satisfaction estimator (refer to \S\ref{sec:basees}). Recall that the base estimator leverages only the dialogue context for USE. The results on SGD and ReDial are shown in Figure~\ref{fig:base_asap}. For SGD, we report the results of both single-task learning and multi-task learning. From Figure~\ref{fig:base_asap}, it can be observed that ASAP consistently outperforms the base estimator over all four metrics on both datasets. This observation validates the effectiveness of the Hawkes process integration module.


\subsection{Contribution of Satisfaction Sequence to Intensity Function}

\begin{figure*}[!t]
   \begin{minipage}[t]{0.312\textwidth}
     \centering
     \includegraphics[width=1.0\linewidth]{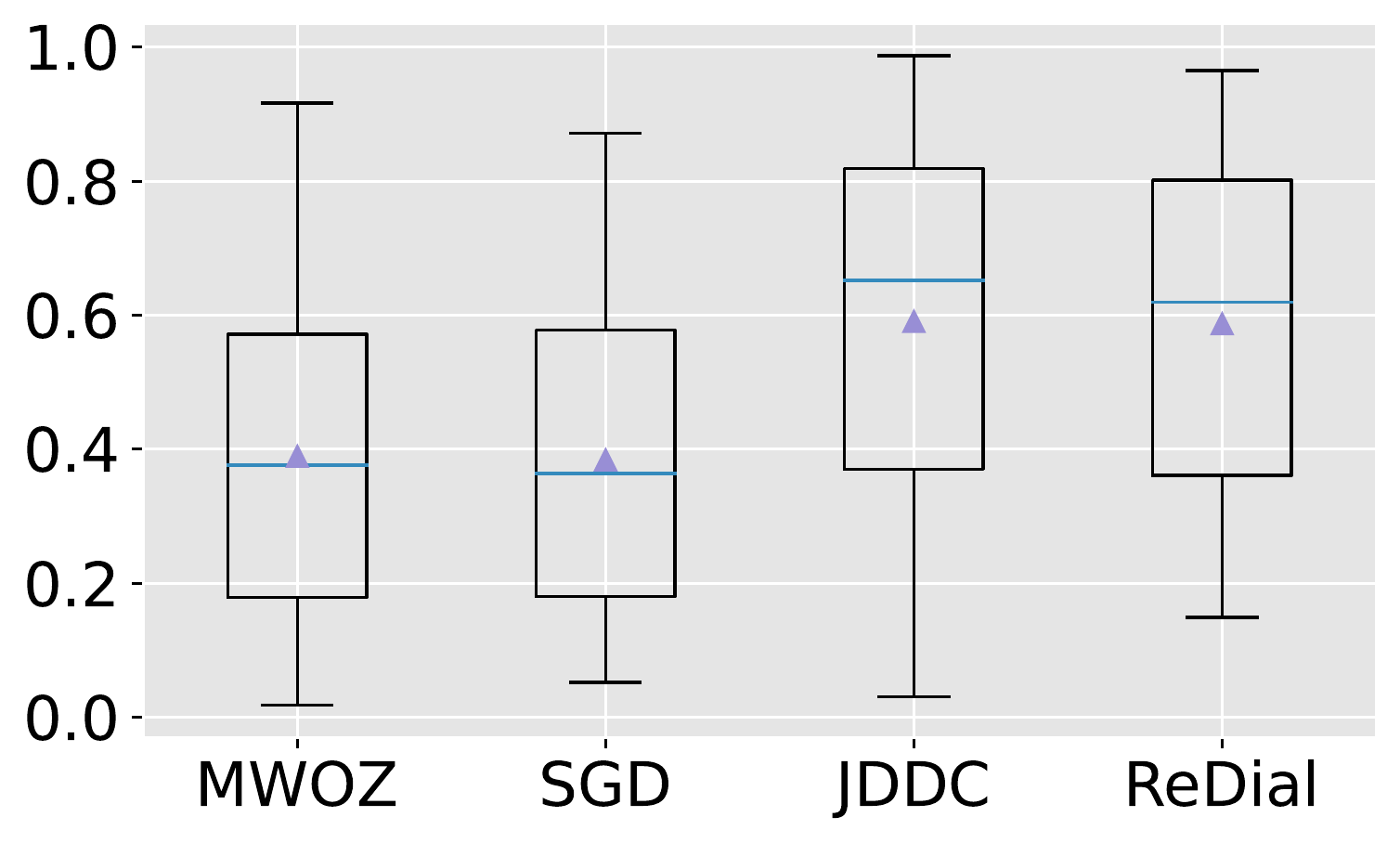}
     \caption{Contribution of satisfaction sequence to intensity function.}\label{fig:alphaalpha}
   \end{minipage}\hfill
   \begin{minipage}[t]{0.322\textwidth}
     \centering
     \includegraphics[width=1.0\linewidth]{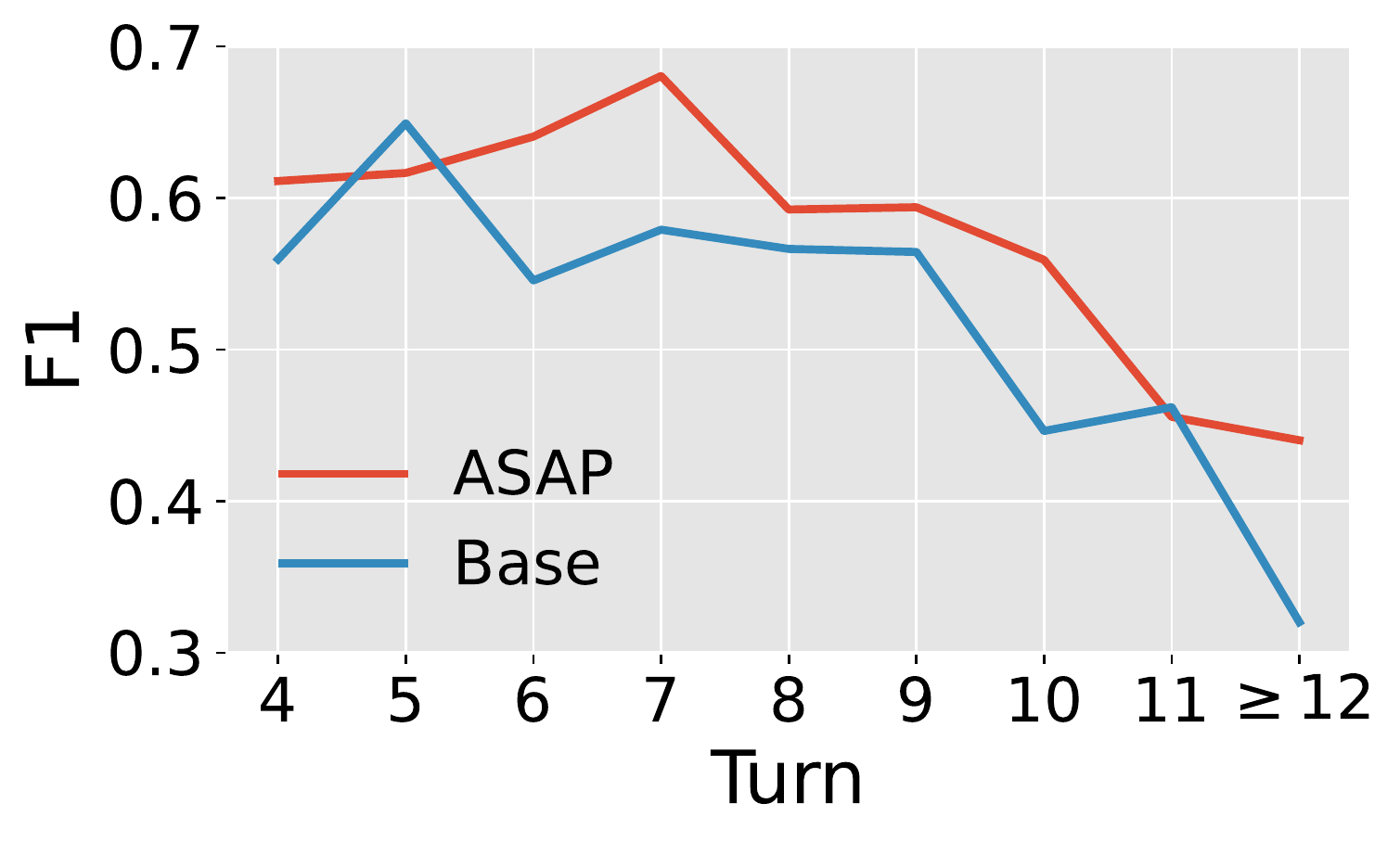}
     \caption{Performance over dialogue turn on ReDial.}\label{fig:turnturn}
   \end{minipage}\hfill
   \begin{minipage}[t]{0.322\textwidth}
     \centering
     \includegraphics[width=1.0\linewidth]{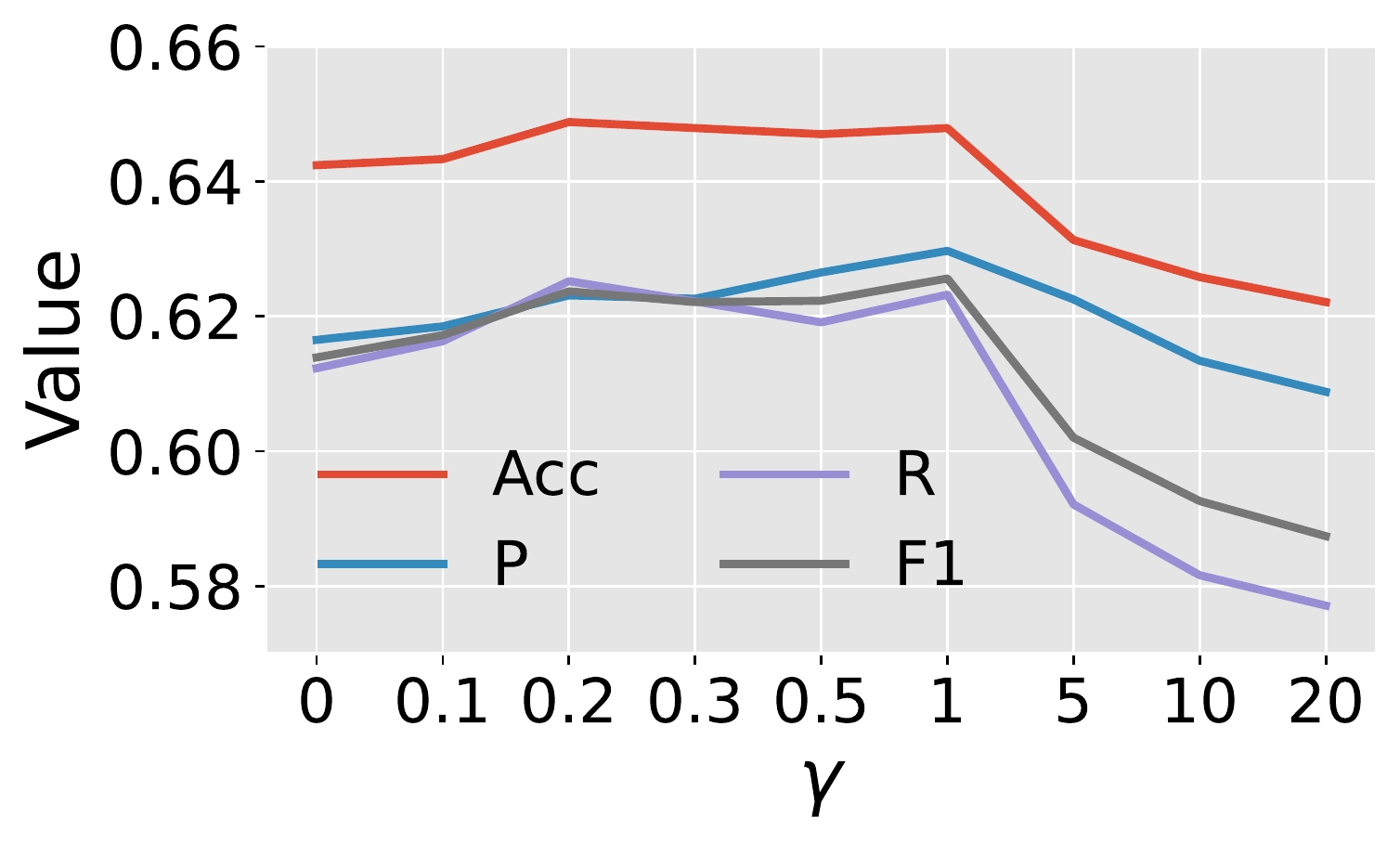}
     \caption{Effects of the parameter $\gamma$ on SGD.}\label{fig:quan}
   \end{minipage}
\end{figure*}

As shown in Eq.~\eqref{eq:kkkkkkkk}, the dialogue context and satisfaction sequence both contribute to the intensity function of the Hawkes process. Here, we explore how much contribution should be attributed to the satisfaction sequence. This study is a supplement to the analysis in the previous section and can provide more insights into the effectiveness of satisfaction dynamics modeling. Considering that the softplus function is monotonically increasing, we can measure the importance of the satisfaction sequence by the value $\exp({\mathtt{MLP}_{s_t}(\bm{x}_t)})/ (\exp({\mathtt{MLP}_{s_t}(\bm{x}_t)}) + \exp({\mathtt{MLP}_{s_t}(\bm{c}_t)}))$. The larger this value is, the more the satisfaction sequence contributes. We calculate this value for all samples in the test set and employ a box plot to show the distribution of these values. The detailed results are provided in Figure \ref{fig:alphaalpha}, where the triangle marker indicates the mean value. We see that the importance of the satisfaction sequence depends on the dataset. For MWOZ and SGD, the dialogue context tends to contribute more than the satisfaction sequence. In contrast, for JDDC and ReDial, the satisfaction sequence tends to be more important. Despite the variance across datasets, we can conclude that the satisfaction sequence generally plays a critical role.

\subsection{Performance over Dialogue Turn}

Given that longer dialogues tend to be more challenging, we further investigate the relationship between the depth of dialogue and the performance of our method. Specifically, we study how the performance changes over dialogue turn. The results of ASAP on ReDial are illustrated in Figure~\ref{fig:turnturn}, where we also report the results of the base estimator for comparison. We omit the results of the first three turns because of their short dialogue context. From Figure~\ref{fig:turnturn}, it can be seen that ASAP outperforms the base estimator in most turns, which again verifies the effectiveness of the Hawkes process integration module. However, we observe that the performance of ASAP and the base estimator degrades when the dialogue is deep. Nonetheless, the performance of ASAP is more robust to the increase of dialogue depth, which should be attributed to the modeling of user satisfaction dynamics.

\subsection{Effects of Parameter $\gamma$}

Figure~\ref{fig:quan} shows the impacts of the parameter $\gamma$ on the performance of our method in the multi-task learning setting. Note that $\gamma$ is used to adjust the weight of the UAR task. From Figure~\ref{fig:quan}, we observe that when $\gamma$
takes small values, the performance is relatively stable. However, the performance drops drastically when $\gamma$ becomes large. This is because when $\gamma$ takes large values, the training objective is dominated by the UAR task. As a consequence, our method fails to optimize the satisfaction estimator.

\section{Related Work}

We briefly review related work on user satisfaction estimation and Hawkes process.

\noindent \textbf{User Satisfaction Estimation.} Evaluation is crucial for the development of dialogue systems \citep{sun2021simulating}. However, evaluating a dialogue system comprehensively can prove to be challenging due to the lack of a clear definition of what constitutes a high-quality dialogue \citep{deriu2021survey}. Typically, a user study is carried out to collect feedback from end users. However, human evaluation is costly and time-intensive.

Another line of approaches is to perform evaluation from the language point of view. The main objective is to measure how natural and syntactically and semantically correct the system responses are \citep{kachuee-etal-2021-self}. For example, several machine translation metrics such as BLEU \citep{papineni-etal-2002-bleu} and ROUGE \citep{lin-2004-rouge} can be used to measure if system responses are consistent with a set of provided answers. These approaches, albeit efficient,  suffer from misalignment with human judgment \citep{novikova-etal-2017-need}.

More recently, user satisfaction estimation has been proposed as an alternative \cite{liang-etal-2021-turn, bodigutla2019multi, sun2021simulating, deng2022user, pan-etal-2022-user}. It leverages human annotations regarding turn-level satisfaction to train an estimator. The estimator is then utilized to perform automatic evaluation by simulating users. Due to this, the evaluation quality depends heavily on the performance of the estimator. In the literature, different approaches have been proposed to train robust estimators \citep{jiang2015automatic, choi2019offline, park2020large, deriu2021survey, deng2022user}. However, none of them considered satisfaction dynamics, which we have shown is a severe deficiency in fully simulating users.

\noindent \textbf{Hawkes Process.}
Hawkes process \citep{hawkes2018hawkes} is a self-exciting process and has been widely used to model sequential data \cite{salehi2019learning}. To enhance the capacity of the standard Hawkes process, several RNNs-based and Transformer-based variants have been proposed \citep{xiao2017modeling, zhang2020self, zuo2020transformer}. All these Hawkes processes are continuous over time. There are also studies on discrete Hawkes processes \citep{seol2015limit, browning2021simple}. However, these discrete versions still predict when the next event happens.

\section{Conclusion}

In this paper, we proposed a new estimator ASAP that adopts the Hawkes process to efficiently capture user satisfaction dynamics across turns within a dialogue. 
Specifically, we devised a discrete version of the continuous Hawkes process to adapt it to the USE task and implemented this discrete version with a Transformer architecture. Extensive experiments on four benchmark datasets demonstrated the superiority of ASAP over baseline USE methods and the effectiveness of the Hawkes process module in modeling user satisfaction dynamics.

\section*{Limitations}
Although our proposed method ASAP is able to outperform baseline estimators, an important factor it ignores is the subjectivity of user satisfaction. In practice, different users may have different degrees of satisfaction with the same dialogue. This implies that ASAP may be effective for some users, but it may also fail to predict true satisfaction for others. In order to adequately simulate a user, it is essential to take the issue of subjectivity into account. Given this, we would like to extend ASAP for personalized satisfaction estimation by incorporating user profile information in the future.

\section*{Acknowledgements}

This work was funded by the EPSRC Fellowship titled “Task Based Information Retrieval” (grant reference number EP/P024289/1) and the Alan Turing Institute.

\bibliography{anthology,custom}

\begin{thebibliography}{48}
\expandafter\ifx\csname natexlab\endcsname\relax\def\natexlab#1{#1}\fi

\bibitem[{Ba et~al.(2016)Ba, Kiros, and Hinton}]{ba2016layer}
Jimmy~Lei Ba, Jamie~Ryan Kiros, and Geoffrey~E Hinton. 2016.
\newblock Layer normalization.
\newblock \emph{arXiv preprint arXiv:1607.06450}.

\bibitem[{Bacry et~al.(2015)Bacry, Mastromatteo, and Muzy}]{bacry2015hawkes}
Emmanuel Bacry, Iacopo Mastromatteo, and Jean-Fran{\c{c}}ois Muzy. 2015.
\newblock Hawkes processes in finance.
\newblock \emph{Market Microstructure and Liquidity}, 1(01):1550005.

\bibitem[{Bodigutla et~al.(2019)Bodigutla, Polymenakos, and
  Matsoukas}]{bodigutla2019multi}
Praveen~Kumar Bodigutla, Lazaros Polymenakos, and Spyros Matsoukas. 2019.
\newblock Multi-domain conversation quality evaluation via user satisfaction
  estimation.
\newblock \emph{arXiv preprint arXiv:1911.08567}.

\bibitem[{Bodigutla et~al.(2020)Bodigutla, Tiwari, Matsoukas, Valls-Vargas, and
  Polymenakos}]{bodigutla-etal-2020-joint}
Praveen~Kumar Bodigutla, Aditya Tiwari, Spyros Matsoukas, Josep Valls-Vargas,
  and Lazaros Polymenakos. 2020.
\newblock \href {https://doi.org/10.18653/v1/2020.findings-emnlp.347} {Joint
  turn and dialogue level user satisfaction estimation on multi-domain
  conversations}.
\newblock In \emph{Findings of the Association for Computational Linguistics:
  EMNLP 2020}, pages 3897--3909, Online. Association for Computational
  Linguistics.

\bibitem[{Browning et~al.(2021)Browning, Sulem, Mengersen, Rivoirard, and
  Rousseau}]{browning2021simple}
Raiha Browning, Deborah Sulem, Kerrie Mengersen, Vincent Rivoirard, and Judith
  Rousseau. 2021.
\newblock Simple discrete-time self-exciting models can describe complex
  dynamic processes: A case study of covid-19.
\newblock \emph{PloS one}, 16(4):e0250015.

\bibitem[{Cai and Chen(2020)}]{cai2020predicting}
Wanling Cai and Li~Chen. 2020.
\newblock Predicting user intents and satisfaction with dialogue-based
  conversational recommendations.
\newblock In \emph{Proceedings of the 28th ACM Conference on User Modeling,
  Adaptation and Personalization}, pages 33--42.

\bibitem[{Cerisara et~al.(2018)Cerisara, Jafaritazehjani, Oluokun, and
  Le}]{cerisara-etal-2018-multi}
Christophe Cerisara, Somayeh Jafaritazehjani, Adedayo Oluokun, and Hoa~T. Le.
  2018.
\newblock \href {https://aclanthology.org/C18-1063} {Multi-task dialog act and
  sentiment recognition on mastodon}.
\newblock In \emph{Proceedings of the 27th International Conference on
  Computational Linguistics}, pages 745--754, Santa Fe, New Mexico, USA.
  Association for Computational Linguistics.

\bibitem[{Chen et~al.(2020)Chen, Liu, Shen, Yuan, Zhou, Wu, He, and
  Zhou}]{chen-etal-2020-jddc}
Meng Chen, Ruixue Liu, Lei Shen, Shaozu Yuan, Jingyan Zhou, Youzheng Wu,
  Xiaodong He, and Bowen Zhou. 2020.
\newblock \href {https://aclanthology.org/2020.lrec-1.58} {The {JDDC} corpus: A
  large-scale multi-turn {C}hinese dialogue dataset for {E}-commerce customer
  service}.
\newblock In \emph{Proceedings of the Twelfth Language Resources and Evaluation
  Conference}, pages 459--466, Marseille, France. European Language Resources
  Association.

\bibitem[{Cho et~al.(2014)Cho, van Merri{\"e}nboer, Bahdanau, and
  Bengio}]{cho-etal-2014-properties}
Kyunghyun Cho, Bart van Merri{\"e}nboer, Dzmitry Bahdanau, and Yoshua Bengio.
  2014.
\newblock \href {https://doi.org/10.3115/v1/W14-4012} {On the properties of
  neural machine translation: Encoder{--}decoder approaches}.
\newblock In \emph{Proceedings of {SSST}-8, Eighth Workshop on Syntax,
  Semantics and Structure in Statistical Translation}, pages 103--111, Doha,
  Qatar. Association for Computational Linguistics.

\bibitem[{Choi et~al.(2019)Choi, Ahmadvand, and Agichtein}]{choi2019offline}
Jason~Ingyu Choi, Ali Ahmadvand, and Eugene Agichtein. 2019.
\newblock Offline and online satisfaction prediction in open-domain
  conversational systems.
\newblock In \emph{Proceedings of the 28th ACM International Conference on
  Information and Knowledge Management}, pages 1281--1290.

\bibitem[{Deng and Lin(2022)}]{deng2022benefits}
Jianyang Deng and Yijia Lin. 2022.
\newblock The benefits and challenges of chatgpt: An overview.
\newblock \emph{Frontiers in Computing and Intelligent Systems}, 2(2):81--83.

\bibitem[{Deng et~al.(2022)Deng, Zhang, Lam, Cheng, and Meng}]{deng2022user}
Yang Deng, Wenxuan Zhang, Wai Lam, Hong Cheng, and Helen Meng. 2022.
\newblock User satisfaction estimation with sequential dialogue act modeling in
  goal-oriented conversational systems.
\newblock In \emph{Proceedings of the ACM Web Conference 2022}, pages
  2998--3008.

\bibitem[{Deriu et~al.(2021)Deriu, Rodrigo, Otegi, Echegoyen, Rosset, Agirre,
  and Cieliebak}]{deriu2021survey}
Jan Deriu, Alvaro Rodrigo, Arantxa Otegi, Guillermo Echegoyen, Sophie Rosset,
  Eneko Agirre, and Mark Cieliebak. 2021.
\newblock Survey on evaluation methods for dialogue systems.
\newblock \emph{Artificial Intelligence Review}, 54(1):755--810.

\bibitem[{Devlin et~al.(2019)Devlin, Chang, Lee, and
  Toutanova}]{devlin-etal-2019-bert}
Jacob Devlin, Ming-Wei Chang, Kenton Lee, and Kristina Toutanova. 2019.
\newblock \href {https://doi.org/10.18653/v1/N19-1423} {{BERT}: Pre-training of
  deep bidirectional transformers for language understanding}.
\newblock In \emph{Proceedings of the 2019 Conference of the North {A}merican
  Chapter of the Association for Computational Linguistics: Human Language
  Technologies, Volume 1 (Long and Short Papers)}, pages 4171--4186,
  Minneapolis, Minnesota. Association for Computational Linguistics.

\bibitem[{Eric et~al.(2020)Eric, Goel, Paul, Sethi, Agarwal, Gao, Kumar, Goyal,
  Ku, and Hakkani-Tur}]{eric-etal-2020-multiwoz}
Mihail Eric, Rahul Goel, Shachi Paul, Abhishek Sethi, Sanchit Agarwal, Shuyang
  Gao, Adarsh Kumar, Anuj Goyal, Peter Ku, and Dilek Hakkani-Tur. 2020.
\newblock \href {https://aclanthology.org/2020.lrec-1.53} {{M}ulti{WOZ} 2.1: A
  consolidated multi-domain dialogue dataset with state corrections and state
  tracking baselines}.
\newblock In \emph{Proceedings of the Twelfth Language Resources and Evaluation
  Conference}, pages 422--428, Marseille, France. European Language Resources
  Association.

\bibitem[{Fu et~al.(2022)Fu, Gao, Zhao, Wen, and Yan}]{fu2022learning}
Tingchen Fu, Shen Gao, Xueliang Zhao, Ji-rong Wen, and Rui Yan. 2022.
\newblock Learning towards conversational ai: A survey.
\newblock \emph{AI Open}, 3:14--28.

\bibitem[{Hancock et~al.(2019)Hancock, Bordes, Mazare, and
  Weston}]{hancock-etal-2019-learning}
Braden Hancock, Antoine Bordes, Pierre-Emmanuel Mazare, and Jason Weston. 2019.
\newblock \href {https://doi.org/10.18653/v1/P19-1358} {Learning from dialogue
  after deployment: Feed yourself, chatbot!}
\newblock In \emph{Proceedings of the 57th Annual Meeting of the Association
  for Computational Linguistics}, pages 3667--3684, Florence, Italy.
  Association for Computational Linguistics.

\bibitem[{Hawkes(2018)}]{hawkes2018hawkes}
Alan~G Hawkes. 2018.
\newblock Hawkes processes and their applications to finance: a review.
\newblock \emph{Quantitative Finance}, 18(2):193--198.

\bibitem[{Hochreiter and Schmidhuber(1996)}]{hochreiter1996lstm}
Sepp Hochreiter and J{\"u}rgen Schmidhuber. 1996.
\newblock Lstm can solve hard long time lag problems.
\newblock \emph{Advances in neural information processing systems}, 9.

\bibitem[{Jiang et~al.(2015)Jiang, Hassan~Awadallah, Jones, Ozertem, Zitouni,
  Gurunath~Kulkarni, and Khan}]{jiang2015automatic}
Jiepu Jiang, Ahmed Hassan~Awadallah, Rosie Jones, Umut Ozertem, Imed Zitouni,
  Ranjitha Gurunath~Kulkarni, and Omar~Zia Khan. 2015.
\newblock Automatic online evaluation of intelligent assistants.
\newblock In \emph{Proceedings of the 24th International Conference on World
  Wide Web}, pages 506--516.

\bibitem[{Jiao et~al.(2019)Jiao, Yang, King, and Lyu}]{jiao-etal-2019-higru}
Wenxiang Jiao, Haiqin Yang, Irwin King, and Michael~R. Lyu. 2019.
\newblock \href {https://doi.org/10.18653/v1/N19-1037} {{H}i{GRU}:
  {H}ierarchical gated recurrent units for utterance-level emotion
  recognition}.
\newblock In \emph{Proceedings of the 2019 Conference of the North {A}merican
  Chapter of the Association for Computational Linguistics: Human Language
  Technologies, Volume 1 (Long and Short Papers)}, pages 397--406, Minneapolis,
  Minnesota. Association for Computational Linguistics.

\bibitem[{Kachuee et~al.(2021)Kachuee, Yuan, Kim, and
  Lee}]{kachuee-etal-2021-self}
Mohammad Kachuee, Hao Yuan, Young-Bum Kim, and Sungjin Lee. 2021.
\newblock \href {https://doi.org/10.18653/v1/2021.naacl-main.319}
  {Self-supervised contrastive learning for efficient user satisfaction
  prediction in conversational agents}.
\newblock In \emph{Proceedings of the 2021 Conference of the North American
  Chapter of the Association for Computational Linguistics: Human Language
  Technologies}, pages 4053--4064, Online. Association for Computational
  Linguistics.

\bibitem[{Lafferty et~al.(2001)Lafferty, McCallum, and
  Pereira}]{lafferty2001conditional}
John~D Lafferty, Andrew McCallum, and Fernando~CN Pereira. 2001.
\newblock Conditional random fields: Probabilistic models for segmenting and
  labeling sequence data.
\newblock In \emph{Proceedings of the Eighteenth International Conference on
  Machine Learning}, pages 282--289.

\bibitem[{Li et~al.(2018)Li, Ebrahimi~Kahou, Schulz, Michalski, Charlin, and
  Pal}]{li2018towards}
Raymond Li, Samira Ebrahimi~Kahou, Hannes Schulz, Vincent Michalski, Laurent
  Charlin, and Chris Pal. 2018.
\newblock Towards deep conversational recommendations.
\newblock \emph{Advances in neural information processing systems}, 31.

\bibitem[{Liang et~al.(2021)Liang, Takanobu, Li, Zhang, Chen, and
  Huang}]{liang-etal-2021-turn}
Runze Liang, Ryuichi Takanobu, Feng-Lin Li, Ji~Zhang, Haiqing Chen, and Minlie
  Huang. 2021.
\newblock \href {https://doi.org/10.18653/v1/2021.ecnlp-1.4} {Turn-level user
  satisfaction estimation in {E}-commerce customer service}.
\newblock In \emph{Proceedings of the 4th Workshop on e-Commerce and NLP},
  pages 26--32, Online. Association for Computational Linguistics.

\bibitem[{Lin(2004)}]{lin-2004-rouge}
Chin-Yew Lin. 2004.
\newblock \href {https://aclanthology.org/W04-1013} {{ROUGE}: A package for
  automatic evaluation of summaries}.
\newblock In \emph{Text Summarization Branches Out}, pages 74--81, Barcelona,
  Spain. Association for Computational Linguistics.

\bibitem[{Loshchilov and Hutter(2017)}]{loshchilov2017decoupled}
Ilya Loshchilov and Frank Hutter. 2017.
\newblock Decoupled weight decay regularization.
\newblock \emph{arXiv preprint arXiv:1711.05101}.

\bibitem[{Mei and Eisner(2017)}]{mei2017neural}
Hongyuan Mei and Jason~M Eisner. 2017.
\newblock The neural hawkes process: A neurally self-modulating multivariate
  point process.
\newblock \emph{Advances in neural information processing systems}, 30.

\bibitem[{Ni et~al.(2022)Ni, Young, Pandelea, Xue, and Cambria}]{ni2022recent}
Jinjie Ni, Tom Young, Vlad Pandelea, Fuzhao Xue, and Erik Cambria. 2022.
\newblock Recent advances in deep learning based dialogue systems: A systematic
  survey.
\newblock \emph{Artificial intelligence review}, pages 1--101.

\bibitem[{Novikova et~al.(2017)Novikova, Du{\v{s}}ek, Cercas~Curry, and
  Rieser}]{novikova-etal-2017-need}
Jekaterina Novikova, Ond{\v{r}}ej Du{\v{s}}ek, Amanda Cercas~Curry, and Verena
  Rieser. 2017.
\newblock \href {https://doi.org/10.18653/v1/D17-1238} {Why we need new
  evaluation metrics for {NLG}}.
\newblock In \emph{Proceedings of the 2017 Conference on Empirical Methods in
  Natural Language Processing}, pages 2241--2252, Copenhagen, Denmark.
  Association for Computational Linguistics.

\bibitem[{Pan et~al.(2022)Pan, Ma, Pflugfelder, and Groh}]{pan-etal-2022-user}
Yan Pan, Mingyang Ma, Bernhard Pflugfelder, and Georg Groh. 2022.
\newblock \href {https://aclanthology.org/2022.sigdial-1.59} {User satisfaction
  modeling with domain adaptation in task-oriented dialogue systems}.
\newblock In \emph{Proceedings of the 23rd Annual Meeting of the Special
  Interest Group on Discourse and Dialogue}, pages 630--636, Edinburgh, UK.
  Association for Computational Linguistics.

\bibitem[{Papineni et~al.(2002)Papineni, Roukos, Ward, and
  Zhu}]{papineni-etal-2002-bleu}
Kishore Papineni, Salim Roukos, Todd Ward, and Wei-Jing Zhu. 2002.
\newblock \href {https://doi.org/10.3115/1073083.1073135} {{B}leu: a method for
  automatic evaluation of machine translation}.
\newblock In \emph{Proceedings of the 40th Annual Meeting of the Association
  for Computational Linguistics}, pages 311--318, Philadelphia, Pennsylvania,
  USA. Association for Computational Linguistics.

\bibitem[{Park et~al.(2020)Park, Yuan, Kim, Zhang, Spyros, Kim, Sarikaya, Guo,
  Ling, Quinn et~al.}]{park2020large}
Dookun Park, Hao Yuan, Dongmin Kim, Yinglei Zhang, Matsoukas Spyros, Young-Bum
  Kim, Ruhi Sarikaya, Edward Guo, Yuan Ling, Kevin Quinn, et~al. 2020.
\newblock Large-scale hybrid approach for predicting user satisfaction with
  conversational agents.
\newblock \emph{arXiv preprint arXiv:2006.07113}.

\bibitem[{Qin et~al.(2021)Qin, Li, Che, Ni, and Liu}]{qin2021co}
Libo Qin, Zhouyang Li, Wanxiang Che, Minheng Ni, and Ting Liu. 2021.
\newblock Co-gat: A co-interactive graph attention network for joint dialog act
  recognition and sentiment classification.
\newblock In \emph{Proceedings of the AAAI Conference on Artificial
  Intelligence}, volume~35, pages 13709--13717.

\bibitem[{Rastogi et~al.(2020)Rastogi, Zang, Sunkara, Gupta, and
  Khaitan}]{rastogi2020towards}
Abhinav Rastogi, Xiaoxue Zang, Srinivas Sunkara, Raghav Gupta, and Pranav
  Khaitan. 2020.
\newblock Towards scalable multi-domain conversational agents: The
  schema-guided dialogue dataset.
\newblock In \emph{Proceedings of the AAAI Conference on Artificial
  Intelligence}, volume~34, pages 8689--8696.

\bibitem[{Rumelhart et~al.(1986)Rumelhart, Hinton, and
  Williams}]{rumelhart1986learning}
David~E Rumelhart, Geoffrey~E Hinton, and Ronald~J Williams. 1986.
\newblock Learning representations by back-propagating errors.
\newblock \emph{nature}, 323(6088):533--536.

\bibitem[{Salehi et~al.(2019)Salehi, Trouleau, Grossglauser, and
  Thiran}]{salehi2019learning}
Farnood Salehi, William Trouleau, Matthias Grossglauser, and Patrick Thiran.
  2019.
\newblock Learning hawkes processes from a handful of events.
\newblock \emph{Advances in Neural Information Processing Systems}, 32.

\bibitem[{Seol(2015)}]{seol2015limit}
Youngsoo Seol. 2015.
\newblock Limit theorems for discrete hawkes processes.
\newblock \emph{Statistics \& Probability Letters}, 99:223--229.

\bibitem[{Song et~al.(2019)Song, Bing, Gao, Lin, Zhao, Wang, Sun, Liu, and
  Zhang}]{song-etal-2019-using}
Kaisong Song, Lidong Bing, Wei Gao, Jun Lin, Lujun Zhao, Jiancheng Wang,
  Changlong Sun, Xiaozhong Liu, and Qiong Zhang. 2019.
\newblock \href {https://doi.org/10.18653/v1/D19-1019} {Using customer service
  dialogues for satisfaction analysis with context-assisted multiple instance
  learning}.
\newblock In \emph{Proceedings of the 2019 Conference on Empirical Methods in
  Natural Language Processing and the 9th International Joint Conference on
  Natural Language Processing (EMNLP-IJCNLP)}, pages 198--207, Hong Kong,
  China. Association for Computational Linguistics.

\bibitem[{Sun et~al.(2021)Sun, Zhang, Balog, Ren, Ren, Chen, and
  de~Rijke}]{sun2021simulating}
Weiwei Sun, Shuo Zhang, Krisztian Balog, Zhaochun Ren, Pengjie Ren, Zhumin
  Chen, and Maarten de~Rijke. 2021.
\newblock Simulating user satisfaction for the evaluation of task-oriented
  dialogue systems.
\newblock In \emph{Proceedings of the 44th International ACM SIGIR Conference
  on Research and Development in Information Retrieval}, pages 2499--2506.

\bibitem[{Vaswani et~al.(2017)Vaswani, Shazeer, Parmar, Uszkoreit, Jones,
  Gomez, Kaiser, and Polosukhin}]{vaswani2017attention}
Ashish Vaswani, Noam Shazeer, Niki Parmar, Jakob Uszkoreit, Llion Jones,
  Aidan~N Gomez, {\L}ukasz Kaiser, and Illia Polosukhin. 2017.
\newblock Attention is all you need.
\newblock \emph{Advances in neural information processing systems}, 30.

\bibitem[{Veli{\v{c}}kovi{\'c} et~al.(2017)Veli{\v{c}}kovi{\'c}, Cucurull,
  Casanova, Romero, Lio, and Bengio}]{velivckovic2017graph}
Petar Veli{\v{c}}kovi{\'c}, Guillem Cucurull, Arantxa Casanova, Adriana Romero,
  Pietro Lio, and Yoshua Bengio. 2017.
\newblock Graph attention networks.
\newblock \emph{arXiv preprint arXiv:1710.10903}.

\bibitem[{Wang et~al.(2018)Wang, Zhang, He, and Zha}]{wang2018supervised}
Lu~Wang, Wei Zhang, Xiaofeng He, and Hongyuan Zha. 2018.
\newblock Supervised reinforcement learning with recurrent neural network for
  dynamic treatment recommendation.
\newblock In \emph{Proceedings of the 24th ACM SIGKDD international conference
  on knowledge discovery \& data mining}, pages 2447--2456.

\bibitem[{Xiao et~al.(2017)Xiao, Yan, Yang, Zha, and Chu}]{xiao2017modeling}
Shuai Xiao, Junchi Yan, Xiaokang Yang, Hongyuan Zha, and Stephen Chu. 2017.
\newblock Modeling the intensity function of point process via recurrent neural
  networks.
\newblock In \emph{Proceedings of the AAAI Conference on Artificial
  Intelligence}, volume~31.

\bibitem[{Yang et~al.(2016)Yang, Yang, Dyer, He, Smola, and
  Hovy}]{yang-etal-2016-hierarchical}
Zichao Yang, Diyi Yang, Chris Dyer, Xiaodong He, Alex Smola, and Eduard Hovy.
  2016.
\newblock \href {https://doi.org/10.18653/v1/N16-1174} {Hierarchical attention
  networks for document classification}.
\newblock In \emph{Proceedings of the 2016 Conference of the North {A}merican
  Chapter of the Association for Computational Linguistics: Human Language
  Technologies}, pages 1480--1489, San Diego, California. Association for
  Computational Linguistics.

\bibitem[{Zhang et~al.(2020)Zhang, Lipani, Kirnap, and Yilmaz}]{zhang2020self}
Qiang Zhang, Aldo Lipani, Omer Kirnap, and Emine Yilmaz. 2020.
\newblock Self-attentive hawkes process.
\newblock In \emph{International conference on machine learning}, pages
  11183--11193. PMLR.

\bibitem[{Zhou et~al.(2022)Zhou, Yang, Rossi, Zhao, and Yu}]{zhou2022neural}
Zihao Zhou, Xingyi Yang, Ryan Rossi, Handong Zhao, and Rose Yu. 2022.
\newblock Neural point process for learning spatiotemporal event dynamics.
\newblock In \emph{Learning for Dynamics and Control Conference}, pages
  777--789. PMLR.

\bibitem[{Zuo et~al.(2020)Zuo, Jiang, Li, Zhao, and Zha}]{zuo2020transformer}
Simiao Zuo, Haoming Jiang, Zichong Li, Tuo Zhao, and Hongyuan Zha. 2020.
\newblock Transformer hawkes process.
\newblock In \emph{International conference on machine learning}, pages
  11692--11702. PMLR.

\end{thebibliography}
\bibliographystyle{acl_natbib}

\clearpage
\appendix

\section{Implementation \& Training Details}

In our experiments, we follow the same procedure as \citet{deng2022user} to pre-process all datasets. For the token-level BERT encoder, we employ the pre-trained BERT-base-uncased model to initialize its weights for MWOZ, SGD, and ReDial. For the JDDC dataset, we use the pre-trained BERT-base-Chinese model for initialization. Both pre-trained models are available from HuggingFace\footnote{\url{https://huggingface.co/docs/transformers/model_doc/bert}}. For the turn-level encoder, we fix the number of attention heads at 12 and set the number of layers (i.e., $L$) to 2. For the score-level encoder (i.e., the Transformer Hawkes process module), we also fix the number of attention heads at 12. But we treat the number of its layers (i.e., $N$) as a hyper-parameter and choose the value from $\{2, 4, 6, 8, 10, 12\}$. The dimension $d$ of the embedding of each satisfaction class is fixed at 768. For both the turn-level encoder and score-level encoder, the hidden size of the Transformer $\mathtt{FFN}$ inner representation layer is set to 3072. All the other involved MLP networks contain only one hidden layer with the hidden size set to 192. The size of their output layers is either the number of satisfaction classes or the number of action types. The ``softness'' parameter $\beta$ of the softplus function is fixed at 1.

AdamW \citep{loshchilov2017decoupled} is exploited as the optimizer, and a linear schedule with warmup is created to adjust the learning rate dynamically. The peak learning rate is chosen from \{1e-5, 2e-5\}. The warmup proportion is set to 0.1. The dropout ratio is also set to 0.1. For all datasets, we train the model for up to 5 epochs. For MWOZ and SGD, we adopt a batch size of 16. While we set the batch size to 24 for ReDial and JDDC. In the multi-task learning setting, we set the parameter $\gamma$ for MWOZ, SGD, and JDDC to 0.5, 1.0, and 0.1, respectively. The best model checkpoints are selected based on the F1 score on the validation set. For all experiments, we use a fixed random seed 42. And it took us around 300 GPU hours to finish the experiments. 

To justify that the performance improvements of our proposed method are significant, we apply the SciPy package's stats.ttest\_rel function\footnote{\url{https://docs.scipy.org/doc/scipy/reference/generated/scipy.stats.ttest_rel.html}} to perform a paired $t$-test against the most competitive baseline USDA and calculate the $p$-value.

\section{Performance of User Action Recognition}

Recall that in the multi-task learning setting, our method ASAP is trained to predict user satisfaction and user action simultaneously. We have presented the results on USE. In this part, we further investigate the performance on the UAR task. The results on MWOZ, SGD, and JDDC are summarized in Table~\ref{tb:uarres}, from which we can see that while ASAP slightly underperforms USDA on MWOZ and SGD according to the official USDA results, its performance is on par with that of USDA based on our reproduced results. Compared to other baselines, ASAP consistently achieves better results on both MWOZ and SGD. However, on the JDDC dataset, we find that the performance of ASAP is relatively low. This is because we have used a small value of 0.1 for $\gamma$ on this dataset. Because of this, during the training phase, ASAP is mainly optimized for the USE task rather than the UAR task. It is worth emphasizing that our focus is on improving the performance of USE instead of UAR in this work. Thus, the reported UAR results are based on the checkpoints which achieve the best USE performance. These checkpoints may not fully demonstrate the capabilities of ASAP on the UAR task. In fact, we empirically found that by setting $\gamma$ to larger values, ASAP can achieve much higher performance on action recognition. But this sacrifices the performance on satisfaction estimation.


\begin{table*}[t]
\centering
\resizebox{\textwidth}{!}{%
\begin{tabular}{lcccccccccccc}
\hline
\multirow{2}{*}{\textbf{Models}} & \multicolumn{4}{c}{\textbf{MWOZ}} & \multicolumn{4}{c}{\textbf{SGD}} & \multicolumn{4}{c}{\textbf{JDDC}} \\ 
\cmidrule(lr){2-5} \cmidrule(lr){6-9} \cmidrule(lr){10-13}
 & \textbf{Acc} & \textbf{P} & \textbf{R} & \textbf{F1} & \textbf{Acc} & \textbf{P} & \textbf{R} & \textbf{F1} & \textbf{Acc} & \textbf{P} & \textbf{R} & \textbf{F1} \\ \hline
JointDAS & 75.1 & 64.5 & 64.7 & 62.8 & 79.5 & 72.1 & 72.7 & 70.9 & 63.4 & 41.8 & 43.6 & 41.1 \\
Co-GAT & 75.6 & 68.5 & 68.4 & 66.6 & 87.5 & 80.9 & 81.5 & 80.2 & 64.2 & 42.5 & 43.6 & 41.5 \\
~~~+BERT & 86.2 & 79.8 & 80.1 & 78.8 & 92.5 & 88.2 & 88.3 & 87.6 & 66.7 & 49.4 & 48.9 & 47.5 \\
JointUSE & 76.5 & 68.7 & 67.7 & 66.9 & 85.0 & 78.0 & 78.9 & 77.3 & 61.8 & 39.0 & 41.8 & 38.8 \\
~~~+BERT & 84.4 & 77.4 & 78.0 & 76.3 & 92.4 & 88.3 & 88.5 & 87.7 & 66.8 & 49.2 & 48.7 & 47.3 \\ \hdashline
USDA & \textbf{87.7} & \textbf{82.8} & \textbf{82.4} & \textbf{81.4} & \textbf{95.8} & \textbf{93.6} & \textbf{93.4} & \textbf{93.1} & \textbf{69.7} & \textbf{53.1} & \textbf{53.0} & \textbf{51.3} \\
USDA$^{\dagger}$ & 86.3 & 81.3 & {\ul 82.2} & 80.3 & {\ul 94.5} & {\ul 91.4} & 91.2 & {\ul 90.8} & {\ul 69.4} & {\ul 52.3} & {\ul 52.1} & {\ul 50.6} \\
\textbf{ASAP} & {\ul 87.0} & {\ul 81.5} & 81.7 & {\ul 80.4} & {\ul 94.5} & 91.2 & {\ul 91.4} & {\ul 90.8} & 47.0 & 19.1 & 26.6 & 20.9 \\ \hline
\end{tabular}%
}
\caption{Comparison of performance on user action recognition. $\dagger$ indicates our reproduced results. The best results are shown in bold and the second-best results are underlined.}
\label{tb:uarres}
\end{table*}

\section{Effects of Number of Layers $N$ in the Score-Level Encoder}

\begin{figure}[h!]
	\centering
	\subfigure[{SGD (single-task)}]{
		\includegraphics[width=0.465\columnwidth]{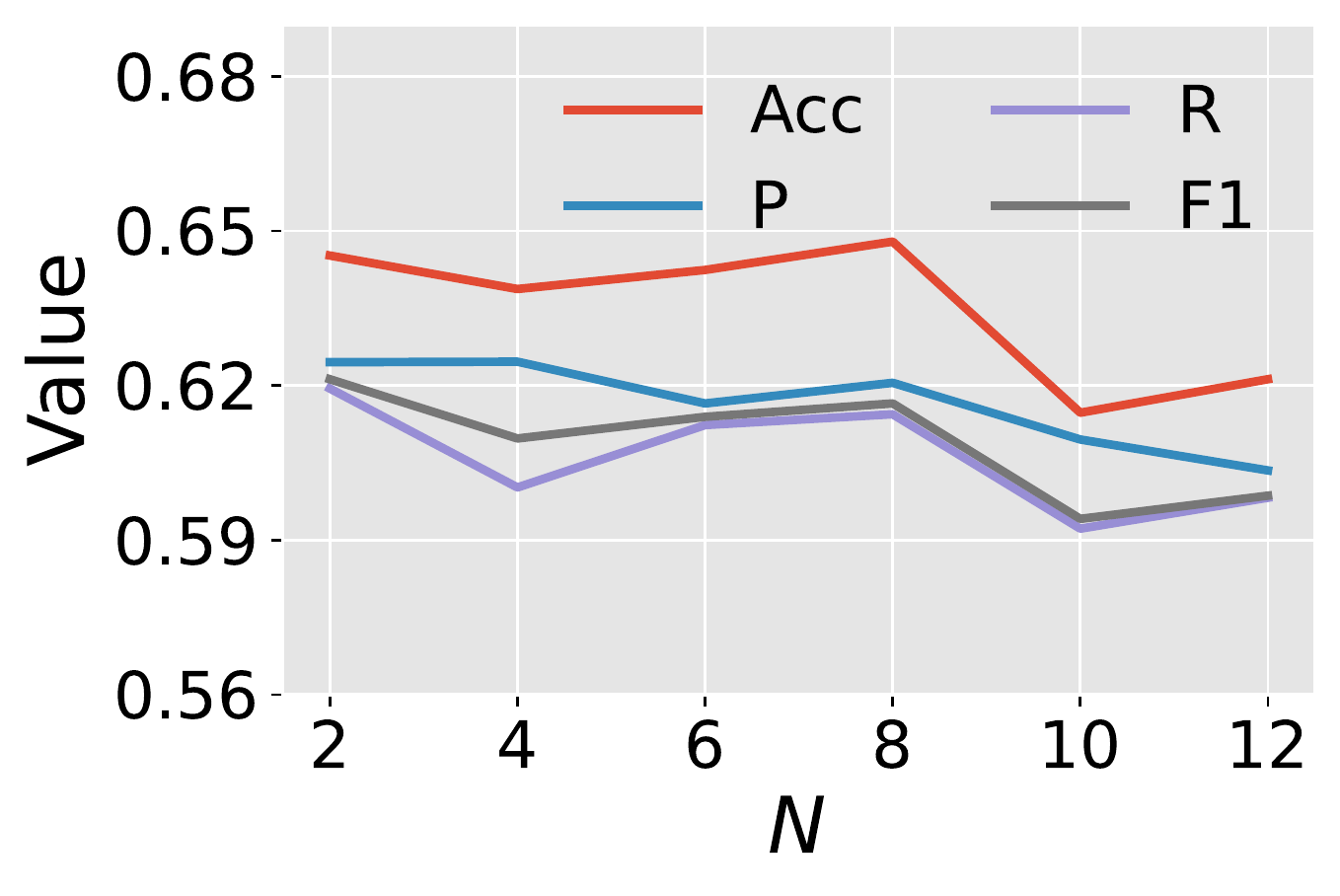}
	} \hfill
	\subfigure[{SGD (multi-task)}]{
		\includegraphics[width=0.465\columnwidth]{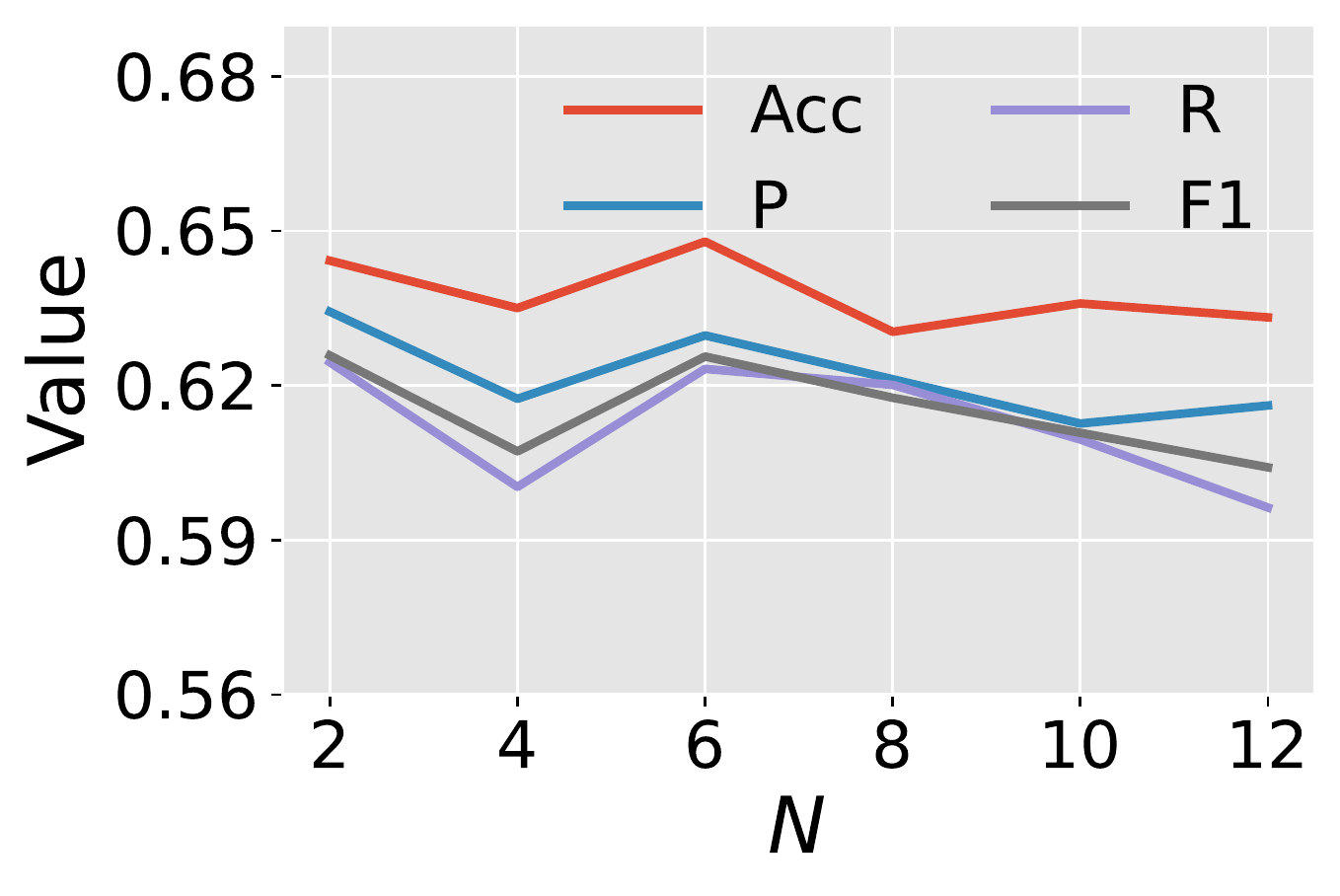}
	}
	\vspace*{-0.2cm}
	\caption{Effects of the number of layers (i.e., $N$) in the score-level encoder on SGD.}
	\label{fig:vary_N}
\end{figure}

\noindent Given that the score-level encoder (i.e., the Transformer Hawkes process module) consists of $N$ layers, it is worth studying the impacts of $N$ on performance by varying its value. For this purpose, we conduct another experiment on the SGD dataset and choose the value of $N$ from  $\{2, 4, 6, 8, 10, 12\}$. We carry out this experiment in both the single-task learning setting and the multi-task learning setting. The results are shown in Figure~\ref{fig:vary_N}. It can be observed that although different values of $N$ lead to different results, the performance is relatively stable. Even so, the performance tends to be higher when $N$ takes smaller values. When $N$ is larger, it is harder to optimize the model because there are more parameters. Additionally, the model is also more prone to overfitting the data.

\end{document}